*Article*

# Semantic Information G Theory and Logical Bayesian Inference for Machine Learning

**Chenguang Lu**

survival99@gmail.com

**Abstract:** An important problem with machine learning is that when label number $n$>2, it is very difficult to construct and optimize a group of learning functions, and we wish that optimized learning functions are still useful when prior distribution $P(x)$ (where $x$ is an instance) is changed. To resolve this problem, the semantic information G theory, Logical Bayesian Inference (LBI), and a group of Channel Matching (CM) algorithms together form a systematic solution. A semantic channel in the G theory consists of a group of truth functions or membership functions. Compared with likelihood functions, Bayesian posteriors, and Logistic functions used by popular methods, membership functions can be more conveniently used as learning functions without the above problem. In Logical Bayesian Inference (LBI), every label's learning is independent. For Multilabel learning, we can directly obtain a group of optimized membership functions from a big enough sample with labels, without preparing different samples for different labels. A group of Channel Matching (CM) algorithms are developed for machine learning. For the Maximum Mutual Information (MMI) classification of three classes with Gaussian distributions on a two-dimensional feature space, 2-3 iterations can make mutual information between three classes and three labels surpass 99% of the MMI for most initial partitions. For mixture models, the Expectation-Maxmization (EM) algorithm is improved and becomes the CM-EM algorithm, which can outperform the EM algorithm when mixture ratios are imbalanced, or local convergence exists. The CM iteration algorithm needs to combine neural networks for MMI classifications on high-dimensional feature spaces. LBI needs further studies for the unification of statistics and logic.

## 1. Introduction

Machine learning needs learning functions and classifiers. In 1922, Fisher [1] proposed the Likelihood Inference (LI) (see Appendix A for all abbreviations in this paper), which uses likelihood functions as learning functions and use the Maximum Likelihood (ML) criterion to optimize learning functions and classifiers. However, when the prior distribution $P(x)$ (where $x$ is an instance) is changed, the optimized likelihood function will be invalid. Since LI cannot make use of prior knowledge, during 1950s, Bayesians proposed Bayesian Inference (BI) [2,3], which uses Bayesian posteriors as learning functions. However, we often only have the prior knowledge of instances instead of labels or model parameters. BI is still not good in these cases. A pair of Logistic (or Sigmoid) functions are often used as learning functions for binary classifications. With a Logistic function and the Bayes' Theorem, we can make use of new $P(x)$ to make new probability prediction for the ML classifier. However, when the labels' number $n$>2, we cannot find proper learning functions that are similar to Logistic functions for Multilabel learning. We call the above problem as "Multilabel-Learning-for-New-$P(x)$ Problem".

Machine learning is to acquire and convey information. The information criterion should be a good criterion. in 1974, Akaike [4] proved that the ML criterion is equal to the minimum Kullback-Leibler (KL) divergence criterion, where KL divergence [5] is also called "KL information". Since then, the information criterion, especially the information criterion that is compatible with the likelihood criterion, has been attracting more researchers' attention [6]. However, KL divergence decreases as likelihood increases, and hence the Least KL divergence is not ideal for the information criterion. Can we use Shannon's mutual information or another information measure for the information criterion?

In 1948, Shannon [7] initiated the classical information theory . In 1949, Weaver [8] proposed the three levels of communication that are relevant to the technical problem resolved by Shannon, the semantic problem relating to meaning and truth, and the effectiveness problem concerning information values. In 1952, Carnap and Bar-Hillel [9] proposed an outline of semantic information theory. Now we can find some different semantic information theories [10-13]. There are also some fuzzy information theories [14-16] and generalized information theories [17,18] relating to semantic information theories. Recently some researchers used Shannon mutual information measure with parameters to optimize neural networks [19,20].

However, before the author, no one brought a learning function with parameters into Shannon's mutual information formula to optimize the learning function and the classifier by a sampling distribution. Therefore, the author tried to develop a semantic information theory that can combine Shannon's information theory and Fisher's likelihood method. The semantic information G theory, or the G theory, is for this purpose. This theory has been developed mainly by the author in the recent three decades [21-26]. It uses membership functions of fuzzy sets proposed by Zadeh [27] as learning functions and treats a membership function as the truth function of a hypothesis. According to Tarski's truth theory [28] and Davidson's truth-conditional semantics [29], the truth function can represent the semantic meaning of a hypothesis.

"The G theory" is used because, in this theory, Semantic Mutual Information (SMI) is a natural generalization of Shannon's Mutual Information (SHMI) ("G" means generalization) so that SHMI is the upper limit of SMI. $G$ also denotes SMI as $D$ denotes average distortion in Shannon's information rate-distortion theory [30]. Replacing $D$ with $G$, the author reformed the rate-distortion function $R(D)$ into the rate- verisimilitude function $R(G)$ [24-25] not only for data compression but also for machine learning.

The G theory has two headstreams: Shannon's information theory and Popper's hypothesis-testing theory [31] (p. 96, 269) [32] (p. 294), which emphasizes that a hypothesis with less logical probability can convey more information if it can survive empirical tests and hence is more preferred.

The semantic information formula proposed by Carnap and Bar-Hillel [9] contains Popper's thought. It is

$$I(p)=\log(1/m_p), \tag{1.1}$$

where $p$ is a hypothesis and $m_p$ is its logical probability. However, this formula does not deal with whether the hypothesis can survive empirical *tests*. Therefore, the author of this paper improved it by Eq. (2.15) in Section 2.1.1. Further, bringing likelihood functions and truth functions into Shannon's mutual information formula, the author obtained the SMI formula.

Cross-entropy is getting popular in machine learning [33]. The G theory uses not only cross-entropy but also mutual cross-entropy [22,25]. The SMI in the G theory is a mutual cross-entropy.

To resolve the Multilabel-Learning-for-New-$P(x)$ problem,, the author found a new inference method, Logical Bayesian Inference (LBI). Bayesians include subjective Bayesians and logical Bayesians. BI was developed by subjective Bayesians, who use subjective probability as the inference tool for statistical inference. Logical Bayesians, such as Keynes and Carnap [34], use logical probability including truth functions as the inference tool for inductive logic. "Logical Bayesian Inference" is used because the (fuzzy) truth function, instead of the Bayesian posterior, is used as the main inference tool. In LBI, both statistical probability and logical probability are used for inference between statistics and logic. BI fits cases with given prior distribution of predictive model $\theta$ whereas LBI fits cases with given prior distribution of instances.

Besides Shannon's information theory and Poppers' hypothesis-testing theory, the G theory is also inherits,, absorbs, or is compatible with

- Fisher's likelihood method for hypothesis-testing [1]
- Zadeh's fuzzy set theory [27,35] for hypotheses' semantic meanings and logical probabilities.
- Carnap and Bar-Hillel's semantic information formula with logical probability [9].
- Floridi's semantic concepts of information [11,36].
- Tarski's truth theory for the definition of truth and logical probability [28].
- Davidson's truth-conditional semantics [29].
- Kullback and Leibler's KL divergence [5].
- Theil's generalized KL formula [36].
- Akaike's proving that the Maximum Likelihood criterion is equal to the Minimum KL divergence criterion [37].
- Donsker-Varadhan representation as a generalized KL formula with Gibs density [38].
- Wittgenstein's thought: meaning lies in uses [39] (p.80).
-
- Bayes' Theorem [40], which can be extended to link likelihood functions and membership functions [41].

Based on the G theory and LBI, the author developed a group of algorithms: Channel Matching (CM) algorithms [41-44], for machine learning. In the CM algorithms, the semantic channel and the Shannon channel mutually match to achieve maximum information (for classifications) or maximum information efficiency ($G/R$) (for mixture models)[1].

These algorithms are used mainly for:

- Making probability predictions with prior knowledge;
- Multilabel learning, belonging to supervised learning；
- The Maximum Mutual Informatioon (MMI) classifications of unseen instances, belonging to semi-supervised learning；
- Mixture models, belonging to unsupervised learning.

Each of them is very difficult and not well resolved. MultilabelMultilabel

The purpose of this paper is to completely introduce the G theory, LBI, and the CM algorithms with background knowledge and applications for readers to fully understand them, especially to understand how to use them to resolve the Multilabel-Learning-for-New-$P(x)$ Problem.

The partial contents of this paper have been introduced in several short papers published in conference proceedings [41-44]. Some contents introduced before are improved in this paper. Such as, one-dimensional examples for the MMI classifications and mixture models now become two-dimensional examples; two formulas for the confirmation measure now become one formula.

According to the author's knowledge, no other one used a semantic information measure to optimize membership functions or truth functions with parameters by sampling distributions; no other one distinguished the statistical probability and the logical probability of a hypothesis and used both in the same formula; no other one proposed the semantic channel with its mathematical representation; no other confirmation measure is compatible with Popper's falsification theory.

This paper also compares the CM algorithms with some popular methods to show their efficiencies.

## 2. Methods I: Background

### 2.1. From the Shannon Information Theory to the Semantic Information G Theory

#### 2.1.1 From Shannon's Mutual Information to Semantic mutual Information

**Definition 1:**

---

[1] The python 3.6 source files for these algorithms producing Figures 9-15 can be found through Appendix B.

- $x$: an instance or data point; $X$: a discrete random variable taking a value $x\in U=\{x_1, x_2, \ldots, x_m\}$.
- $y$: a hypothesis or label; $Y$ is a discrete random variable taking a value $y\in V=\{y_1, y_2, \ldots, y_n\}$.
- $P(y_j|x)$ (with certain $y_j$ and variable $x$): a Transition Probability Function (TPF) (which is so called by Shannon [7]).

Shannon call $P(X)$ as the source, $P(Y)$ as the destination, and $P(Y|X)$ as the channel. A Shannon's channel is a transition probability matrix or a group of transition probability functions:

$$P(Y|X) \Leftrightarrow \begin{bmatrix} P(y_1|x_1) & P(y_1|x_2) & \ldots & P(y_1|x_m) \\ P(y_2|x_1) & P(y_2|x_2) & \ldots & P(y_2|x_m) \\ \ldots & \ldots & \ldots & \ldots \\ P(y_n|x_1) & P(y_n|x_2) & \ldots & P(y_n|x_m) \end{bmatrix} \Leftrightarrow \begin{bmatrix} P(y_j|x) \\ P(y_j|x) \\ \ldots \\ P(y_n|x) \end{bmatrix}, \tag{2.1}$$

where $\Leftrightarrow$ means equivalence. Note TPF $P(y_j|x)$ is not normalized unlike conditional probability function $P(y|x_i)$, in which $y$ is variable, and $x_i$ is constant. We will discuss that the TPF can be used for traditional Bayes' prediction in Section 2.2.1.

Shannon's entropies of $X$ and $Y$ are:

$$H(X) = -\sum_j P(x_i)\log P(x_i), \tag{2.2}$$

$$H(Y) = -\sum_j P(y_j)\log P(y_j). \tag{2.3}$$

Shannon's posterior entropies of $X$ and $Y$ are:

$$H(X|Y) = -\sum_j\sum_i P(x_i, y_j)\log P(x_i|y_j), \tag{2.4}$$

$$H(Y|X) = -\sum_j\sum_i P(x_i, y_j)\log P(y_j|x_i). \tag{2.5}$$

Shannon's mutual information are

$$\begin{aligned} I(X;Y) &= -\sum_j\sum_i P(x_i, y_j)\log\frac{P(x_i|y_j)}{P(x_i)} = H(X) - H(X|Y) \\ &= -\sum_j\sum_i P(x_i, y_j)\log\frac{P(y_j|x_i)}{P(y_j)} = H(Y) - H(Y|X). \end{aligned} \tag{2.6}$$

If $Y=y_j$, mutual information $I(X; Y)$ will become Kullback-Leibler (KL) divergence:

$$I(X;y_j) = \sum_i P(x_i|y_j)\log\frac{P(x_i|y_j)}{P(x_i)} = \sum_i P(x_i|y_j)\log\frac{P(y_j|x_i)}{P(y_j)}. \tag{2.7}$$

Some researchers use the following formula to measure the information between $x_i$ and $y_j$:

$$I(x_i;y_j) = \log\frac{P(x_i|y_j)}{P(x_i)} = \log\frac{P(y_j|x_i)}{P(y_j)}. \tag{2.8}$$

Since $I(x_i; y_j)$ may be negative, Shannon did not use it. Shannon explains that information is the reduced uncertainty or the saved average code word length. The author thinks that the above formula is meaningful because negative information means that a bad prediction may increase the uncertainty or the code word length.

Since Shannon's information theory cannot measure semantic information, Carnap and Bar-Hillel proposed a semantic information formula: $I(p)=\log[1/m_p]$. Since $I(p)$ is irrelative to whether the prediction is correct or not, this formula is unpractical.

Zhong [12] makes use of DeLuca and Termini's fuzzy entropy [14] to define the semantic information measure:

$$I(y_j) = \log 2 + [t_j \log t_j + (1 - t_j) \log(1 - t_j)], \tag{2.9}$$

where $t_j$ is "the logic truth" of $y_j$. However, according to this formula, whatever $t_j$=1 or $t_j$=0, the information reaches its maximum 1 bit. This result is not expected. Therefore, this formula is unreasonable. This problem is also met by other semantic or fuzzy information theories that use DeLuca and Termini's fuzzy entropy [14].

Floridi's semantic information formula [11,36] is a little complicated. It can ensure that the information conveyed by a tautology or a contradiction reaches its minimum 0. However, according to common sense, a wrong prediction or a lie is worse than a tautology. As to how the semantic information is related to the deviation and how the amount of semantic information of a correct prediction is different from that of a wrong prediction, we cannot get clear answers from his formula.

The author proposed an improved semantic information measure in 1990 [21] and developed the G theory later.

According to Tarski's truth theory [28], $P(X \in \theta_j)$ is equivalent to $P$("$X \in \theta$" is true)=$P(y_j$ is true). According to Davidson's truth condition semantics [29], the truth function of $y_j$ ascertains the semantic meaning of $y_j$. Following Tarski and Davidson, we define as follows:

**Definition 2:**

- $\theta_j$ is a fuzzy subset of $U$ and is used to explain the semantic meaning of a predicate $y_j(X)$="$X \in \theta_j$." The $\theta_j$ is a model or a group of model parameters. If $\theta_j$ is non-fuzzy, we may replace it with $A_j$.
- A probability is defined with "=", such as $P(y_j)$=$P(Y$=$y_j)$, is a statistical probability; a probability is defined with "∈", such as $P(X \in \theta_j)$, is a logical probability. To distinguish $P(Y$=$y_j)$ and $P(X \in \theta_j)$, we define $T(\theta_j)$=$P(X \in \theta_j)$ as the logical probability of $y_j$.
- $T(\theta_j|x)$=$P(x \in \theta_j)$=$P(X \in \theta_j|X$=$x)$ is the conditional logical probability function of $y_j$; it is also called the (fuzzy) truth function of $y_j$ or the membership function of $\theta_j$.

A group of TPFs $P(y_j|x)$, $j$=1,2,…,$n$, form a Shannon channel whereas a group of membership functions $T(\theta_j|x)$, $j$=1,2,…$n$, form a semantic channel:

$$T(\theta|X) \Leftrightarrow \begin{bmatrix} T(\theta_1|x_1) & T(\theta_1|x_2) & \dots & T(\theta_1|x_m) \\ T(\theta_2|x_1) & T(\theta_2|x_2) & \dots & T(\theta_2|x_m) \\ \dots & \dots & \dots & \dots \\ T(\theta_n|x_1) & T(\theta_n|x_2) & \dots & T(\theta_n|x_m) \end{bmatrix} \Leftrightarrow \begin{bmatrix} T(\theta_1|x) \\ T(\theta_2|x) \\ \dots \\ T(\theta_n|x) \end{bmatrix}. \tag{2.10}$$

The Shannon channel $P(Y|X)$ and the semantic channel $T(\theta|X)$ are illustrated in Figure 1.

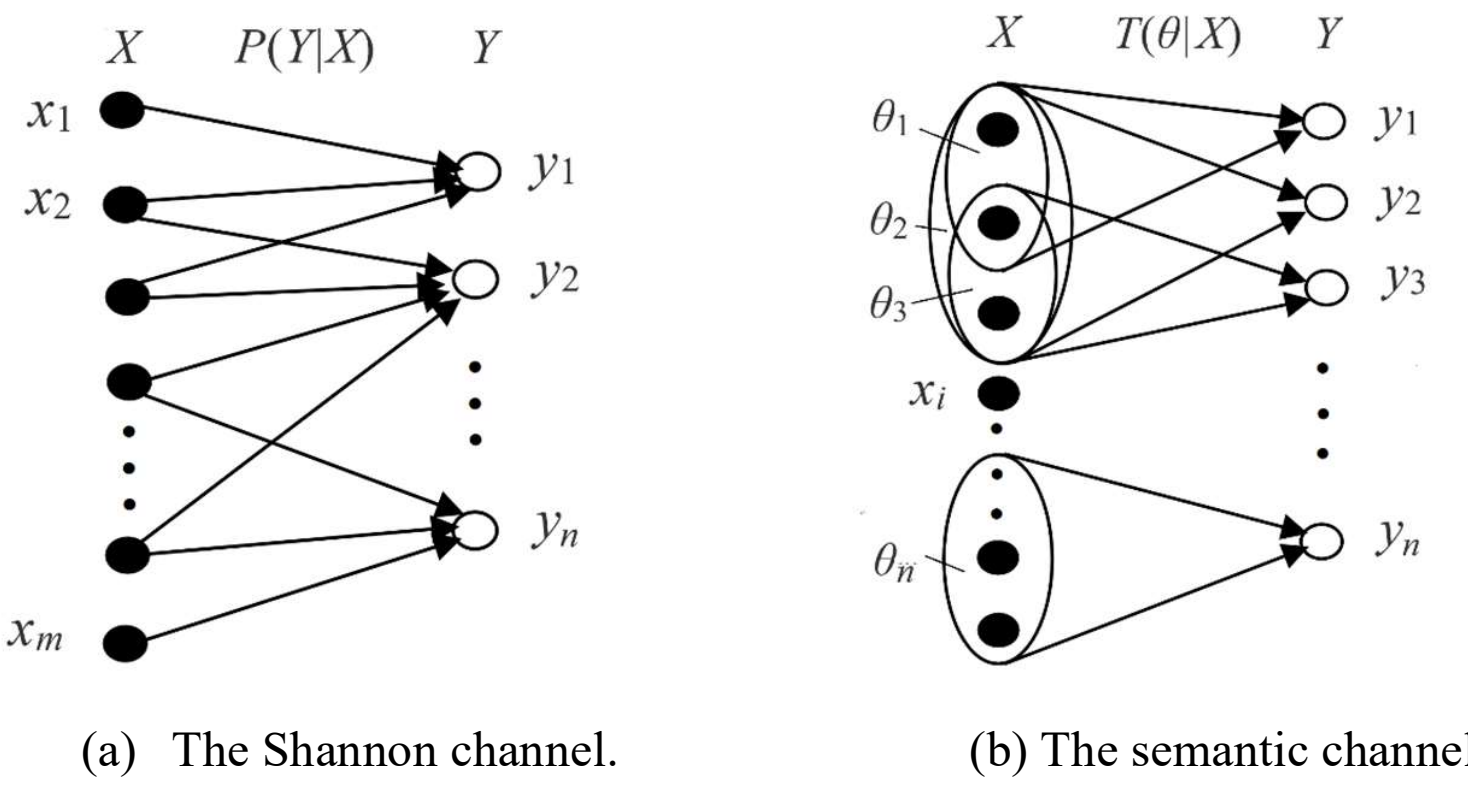

(a) The Shannon channel. (b) The semantic channel

**Figure 1.** The Shannon channel and the semantic channel. The semantic meaning of $y_j$ is ascertained by the membership relation between $x$ and $\theta_j$. A fuzzy set $\theta_j$ may be overlapped or included by another.

The Shannon channel indicates the correlation between $X$ and $Y$ whereas the semantic channel indicates the fuzzy denotations of a set of labels. For the weather forecasts between an observatory and its audience, the Shannon channel indicates the rule by which the observatory selects labels or forecasts, whereas the semantic channel indicates the semantic meanings of these forecasts understood by the audience.

The expectation of the truth function is the logical probability:

$$T(\theta_j)=\sum_i P(x_i)T(\theta_j\mid x_i), \tag{2.11}$$

which was proposed by Zadeh [35] as the probability of a fuzzy event. This logical probability is a little different from $m_p$ defined by Carnap and Bar-Hillel [9]. The latter only rests with a hypothesis' denotation. For example, $y_j$ is a hypothesis: "$X$ is infected by HIV" (HIV: Human Immunodeficiency Virus) or a label "HIV infected". Its logical probability $T(\theta_j)$ is very small for normal people because HIV-infected people are rare. However, $m_p$ is irrelative to $P(x)$; it may be 1/2.

Note that the statistical probability is normalized whereas the logical probability is not. When $\theta_0$, $\theta_1$, …, $\theta_n$ form a cover of $U$, there are $P(y_0)+P(y_1)+\ldots+P(y_n)=1$ and $T(\theta_0)+T(\theta_1)+\ldots+T(\theta_n)\geqslant 1$.

For example, $U$ is a set of different ages. There are nonfuzzy subsets of $U$ (see Figure 2): $A_1$={adults}=$\{x\mid x\geq 18\}$, $A_0$={juveniles}=$\{x\mid x<18\}=A_1^c$ ($^c$ means a complementary set), and $A_2$={young people}=$\{x\mid 15\leq x\leq 35\}$. The three sets form a cover of $U$. There is $T(A_0)+T(A_1)=1$. If $T(A_2)=0.3$; the sum of the three logical probabilities is 1.3>1. However, the sum of three statistical probabilities $P(y_0)+P(y_1)+P(y_2)$ must be less or equal to 1. If $y_2$ is correctly used, $P(y_2)$ will change from 0 to 0.3. If $A_0$, $A_1$, and $A_2$ become fuzzy sets, the conclusion is the same. Consider a tautology "There will be rain or will not be rain tomorrow". Its logical probability is 1 whereas its statistical probability is close to 0 because it is rarely selected.

We can put $T(\theta_j\mid x)$ and $P(x)$ into the Bayes' formula to obtain a likelihood function [21]:

$$P(x\mid\theta_j)=\frac{T(\theta_j\mid x)P(x)}{T(\theta_j)},\ \ T(\theta_j)=\sum_i T(\theta_j\mid x_i)P(x_i), \tag{2.1}$$

$P(x\mid\theta_j)$ can be called the semantic Bayes' prediction or the semantic likelihood function. According to Dubois and Prade's paper [45], Thomas [46] and others proposed similar formulas earlier.

Assume that the maximum of $T(\theta_j\mid x)$ is 1. From $P(x)$ and $P(x\mid\theta_j)$, we can obtain

$$T(\theta_j\mid x)=\frac{T(\theta_j)P(x\mid\theta_j)}{P(x)},\ T(\theta_j)=1/\max[P(x\mid\theta_j)/P(x)], \tag{2.13}$$

where max[.] means the maximum of the function in [.] for different $x$. The author [41] proposed the third type of Bayes' theorem which consists of the above two formulas. This theorem can convert the likelihood function and the membership function or the truth function form one to another when $P(x)$ is given. The Eq. (2.13) is compatible with Wang's fuzzy sets falling shadow theory [41,47].

Figure 2 illustrates the relationship between $P(x\mid\theta_j)$ and $T(\theta_j\mid x)$ for given $P(x)$, where $x$ is an age; label $y_j$="Youth"; $\theta_j$ is a non-fuzzy set and hence becomes $A_j$.

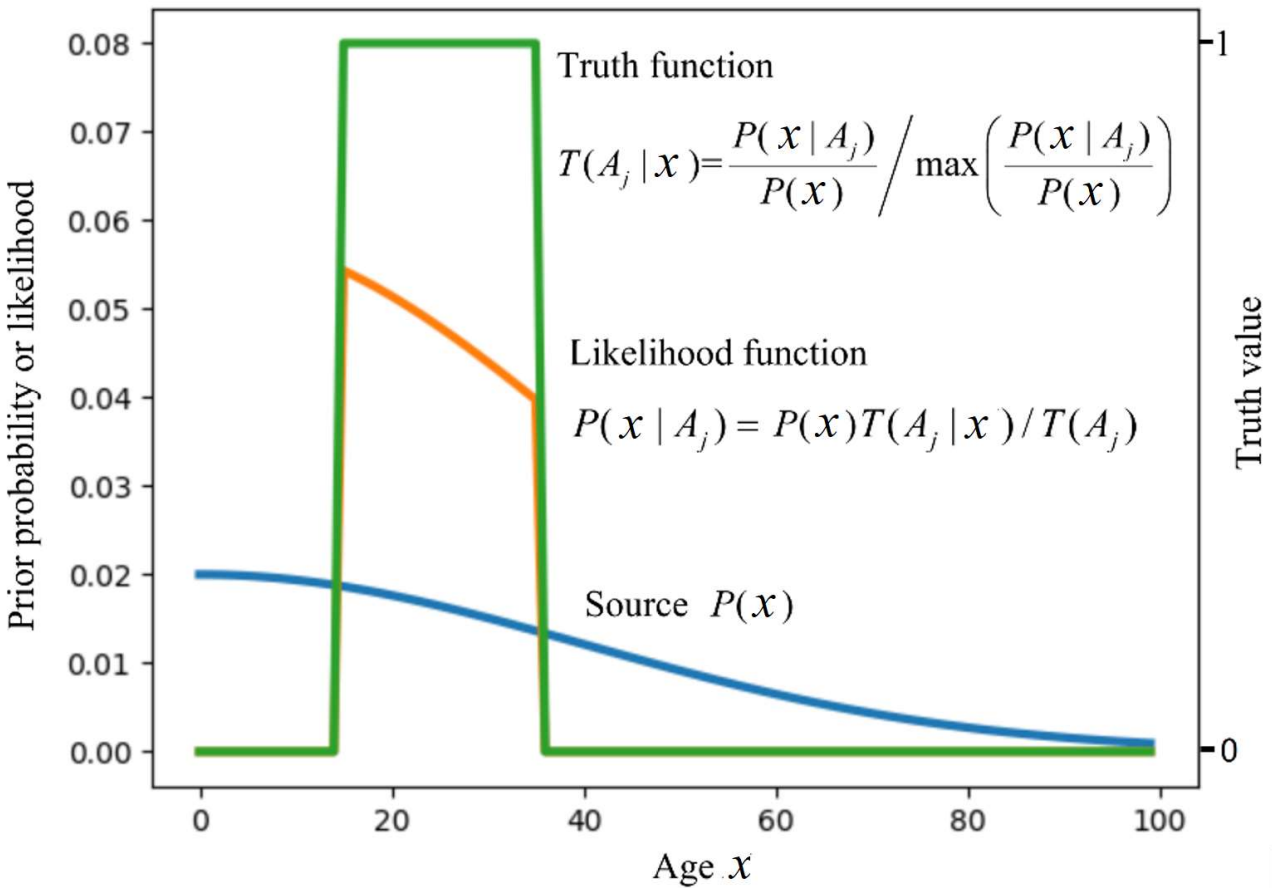

**Figure 2** Relationship between $T(A_j|x)$ and $P(x|A_j)$ for given $P(x)$.

We use Global Positioning System (GPS) as an example to show the semantic Bayes' prediction.

**Example 2.1** A GPS device is used in a train and hence $P(x)$ is uniformly distributed on a line (see Figure 3). The GPS pointer has a deviation. Try to find the most possible position of the GPS device.

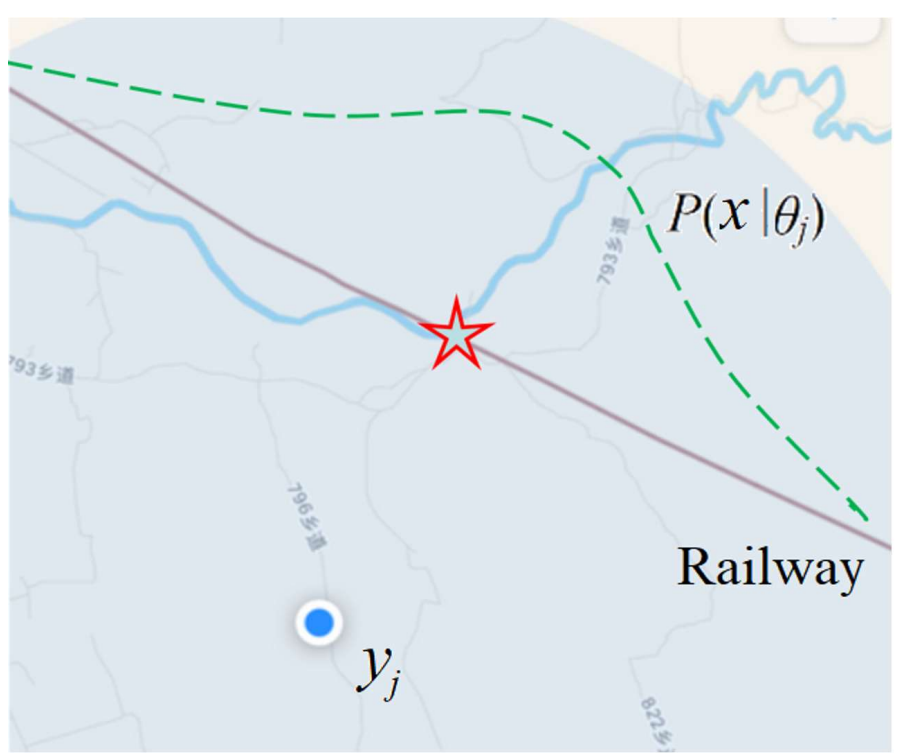

**Figure 3** Illustrating a GPS device's positioning with a deviation. The round point is the pointed position with a deviation and the star is the most possible position.

The semantic meaning of the GPS pointer can be expressed by

$$T(\theta_j|x)=\exp[-(x-x_j)^2/(2\sigma^2)], \tag{2.14}$$

where $x_j$ is the pointed position by $y_j$, and σ is the Root Mean Square (RMS). For simplicity, here we assume $x$ is one-dimensional.

According to Eq. (2.12), we can predict that the star is the most possible position. Most people can make the same prediction without using any mathematical formula. It seems that human brains automatically use a similar method: predicting according to the fuzzy denotation of $y_j$.

In semantic communication, we often see hypotheses or predictions such as "The temperature is about ten ℃", "Time is about seven O'clock", and "The stock index will go up about 10% next month". Every one of them may be represented by $y_j$="X is about $x_j$". We can also express their truth functions by Eq. (2.14).

The author defines the (amount of) semantic information conveyed by $y_j$ about $x_i$ with log-normalized-likelihood :

$$I(x_i;\theta_j)=\log\frac{P(x_i|\theta_j)}{P(x_i)}=\log\frac{T(\theta_j|x_i)}{T(\theta_j)}. \tag{2.15}$$

Bringing Eq. (2.14) into this formula, we have

$$I(x_i;\theta_j) = \log[1/T(\theta_j)] - (x_i - x_j)^2/(2\sigma^2), \tag{2.16}$$

by which we can explain that this information is equal to Carnap-Bar-Hillel information minus relative deviation square. This formula is illustrated in Figure 4.

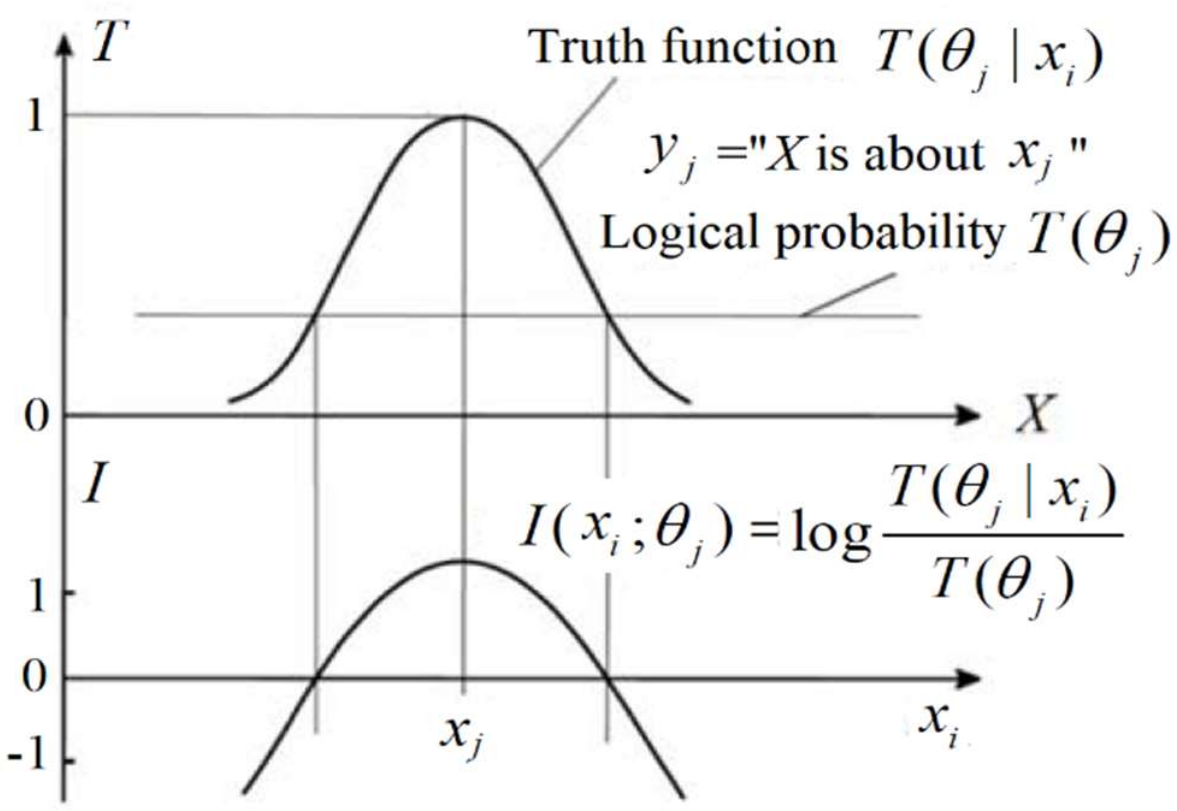

**Figure 4**. The semantic information conveyed by $y_j$ about $x_i$.

Figure 4. indicates that the less the logical probability is, the more information there is; the larger the deviation is, the less information there is; a wrong hypothesis will convey negative information. These conclusions accord with Popper's thought [32] (p. 294).

To average $I(x_i; \theta_j)$, we have generalized KL information

$$I(X;\theta_j) = \sum_i P(x_i \mid y_j) \log \frac{P(x_i|\theta_j)}{P(x_i)} = \sum_i P(x_i \mid y_j) \log \frac{T(\theta_j \mid x_i)}{T(\theta_j)}, \tag{2.17}$$

where $P(x_i|y_j)$ ($i$=1,2,…) is the sampling distribution, which may be unsmooth or discontinuous.

Theil proposed a generalized KL information formula earlier with three probability distributions [37]. However, in Eq. (2.17), $T(\theta_j)$ is a constant. If $T(\theta_j|x)$ is an exponential function with $e$ as the base, then Eq. (2.17) will become the Donsker-Varadhan representation [19,38].

Akaike [4] proved that the Least KL divergence criterion is equivalent to the Maximum likelihood (ML) criterion. We can follow Akaike to prove that the Maximum Semantic Information (MSI) criterion, e.g., maximum generalized KL information criterion, is also equivalent to the ML criterion.

**Definition 3**: **D** is a sample with labels: $\{(x(t), y(t)) | t=1 \text{ to } N; x(t) \in U; y(t) \in V\}$, which includes $n$ different sub-samples or conditional samples $\mathbf{X}_j$, $j$=1, 2, …. Every sub-sample includes data points $x(1)$, $x(2)$, …, $x(N_j) \in U$ with label $y_j$. If $\mathbf{X}_j$ is large enough, we can obtain distribution $P(x|y_j)$ from $\mathbf{X}_j$. If $y_j$ is unknown in $\mathbf{X}_j$, we replace $\mathbf{X}_j$ with $\mathbf{X}$ and $P(x|y_j)$ with $P(x|.)$.

Assume that there are $N_j$ data points in $\mathbf{X}_j$; $N_{ji}$ data points are $x_i$. When $N_j$ data points in $\mathbf{X}_j$ come from Independent and Identically Distributed (IID) random variables, we have the likelihood

$$\begin{aligned}\log P(\mathbf{X}_j \mid \theta_j) &= \log P(x(1), x(2), ..., x(N)|\theta_j) = \log \prod_i P(x_i \mid \theta_j)^{N_{ji}} \\ &= N_j \sum_i P(x_i \mid y_j) \log P(x_i|\theta_j) = -N_j H(X \mid \theta_j).\end{aligned} \tag{2.18}$$

Since

$$I(X;\theta_j)=\sum_i P(x_i\mid y_j)\log\frac{P(x_i\mid\theta_j)}{P(x_i)}=-H(X\mid\theta_j)+\sum_i P(x_i\mid y_j)\log P(x_i), \quad (2.19)$$

$I(X;\theta_j)$ and $\log P(\mathbf{X}_j|\theta_j)$ reach their maxima at the same time as $\theta_j$ changes, and hence the two criterions are equivalent. It is easy to prove that when $P(x|\theta_j)=P(x|y_j)$, $I(X;\theta_j)$ and $\log P(\mathbf{X}_j|\theta_j)$ reach their maxima.

When the sample $\mathbf{X}_j$ is huge, letting $P(x|\theta_j)=P(x|y_j)$, we can obtain the optimized truth function:

$$T^*(\theta_j|x) = [P^*(x|\theta_j)/P(x)]/\max[P^*(x|\theta_j)/P(x)]= [P(x|y_j)/P(x)]/\max[P(x|y_j)/P(x)]. \quad (2.20)$$

We can also obtain

$$T^*(\theta_j|x)= P^*(\theta_j|x)/\max[P^*(\theta_j|x)] =P(y_j|x)/\max[P(y_j|x)]. \quad (2.21)$$

This formula clearly indicates how the semantic channel matches the Shannon channel. It is compatible with Wittgenstein's thought: meaning lies in uses [39] (p.80).

To average $I(X;\theta_j)$ for different $y$, we have the Semantic Mutual Information (SMI) formula

$$\begin{aligned} I(X;\theta) &= \sum_j P(y_j)\sum_i P(x_i\mid y_j)\log\frac{P(x_i|\theta_j)}{P(x_i)} \\ &= \sum_i\sum_j P(x_i)P(y_j\mid x_i)\log\frac{T(\theta_j\mid x_i)}{T(\theta_j)}. \end{aligned} \quad (2.22)$$

If $P(x|\theta_j)=P(x|\mathrm{y}_j)$ or $T(\theta_j|x)\propto P(y_j|x)$ for different $y_j$, which means that the semantical channel matches the Shannon channel, the SMI will be equal to Shannon's Mutual Information (SHMI).

Bringing Eq. (2.13) into the above formula, we have

$$\begin{aligned} I(X;\theta) &= H(\theta)-H(\theta\mid X) \\ &= -\sum_j P(y_j)\log T(\theta_j)-\sum_j\sum_i P(x_i,y_j)(x_i-\mu_j)^2/(2\sigma_j^{\,2}). \end{aligned} \quad (2.23)$$

*It is easy to find that the maximum SMI criterion is a special Regularized Least Square (RLS) criterion [33]. $H(\theta|X)$ is the mean squared error, and $H(\theta)$ is the negative regularization term. The difference is that $H(\theta)$ does not penalize every parameter. It only penalizes parameters for relative deviations. The important is that the maximum SMI criterion is also compatible with the ML criterion.2.1.2. From Rate-Distortion Function $R(D)$ to Rate-Verisimilitude Function $R(G)$*

The $R(G)$ function will be used to explain the convergence of iterative algorithms for the MMI classifications and mixture models.

Shannon proposed rate-distortion function $R(D)$ [30]. $R(G)$ [25] is an extension of $R(D)$. In $R(D)$, $R$ is the information rate, $D$ is the upper limit of average distortion. $R(D)$ means that for given $D$, $R(D)$ is the minimum of SHMI $I(X;Y)$.

The rate distortion function with parameter $s$ [48] (P. 32) includes two formulas:

$$\begin{aligned} D(s) &= \sum_i\sum_j d_{ij}P(x_i)P(y_j)\exp(sd_{ij})/\lambda_i, \\ R(s) &= sD(s)-\sum_i P(x_i)\ln\lambda_i, \end{aligned} \quad (2.24)$$

where $\lambda_i=\sum_j P(y_j)\exp(sd_{ij})$ is the partition function.

Let $d_{ij}$ be replaced with $I_{ij}= I(x_i;y_j)=\log[T(\theta_j|x_i)/T(\theta_j)]= \log[P(x_i|\theta_j)/P(x_i)]$, and let $G$ be the lower limit of $I(X;\theta)$. The information rate for given $G$ and source $P(X)$ is defined as

$$R(G)=\min_{P(Y|X):I(X;\theta)\geq G} I(X;Y). \tag{2.25}$$

Popper [32] proposed to use verisimilitude instead of correctness to evaluate a hypothesis. Verisimilitude includes both correctness and precision. Hence, $I(x_i; \theta_j)$ can be a good measure for the verisimilitude of $y_j$ reflecting $x_i$; we call $R(G)$ as the rate-verisimilitude function.

Following the derivation of $R(D)$ [48] (p. 31), we can obtain

$$\begin{aligned}
&G(s)=\sum_i\sum_j I_{ij}P(x_i)P(y_j)2^{sI_{ij}})/\lambda_i=\sum_i\sum_j I_{ij}P(x_i)P(y_j)m_{ij}^s/\lambda_i,\\
&R(s)=sG(s)-\sum_i P(x_i)\log\lambda_i,\\
&m_{ij}=T(\theta_j\mid x_i)/T(\theta_j),\ \lambda_i=\sum_j P(y_j)m_{ij}^s,
\end{aligned} \tag{2.26}$$

where $m_{ij}=T(\theta_j|x_i)/T(\theta_j)=P(x_i|\theta_j)/P(x_i)$ is the normalized likelihood; $\lambda_i=\sum_j P(y_j)m_{ij}^s$. The shape of any $R(G)$ function is a bowl-like curve with second derivative >0 as shown in Figure 5.

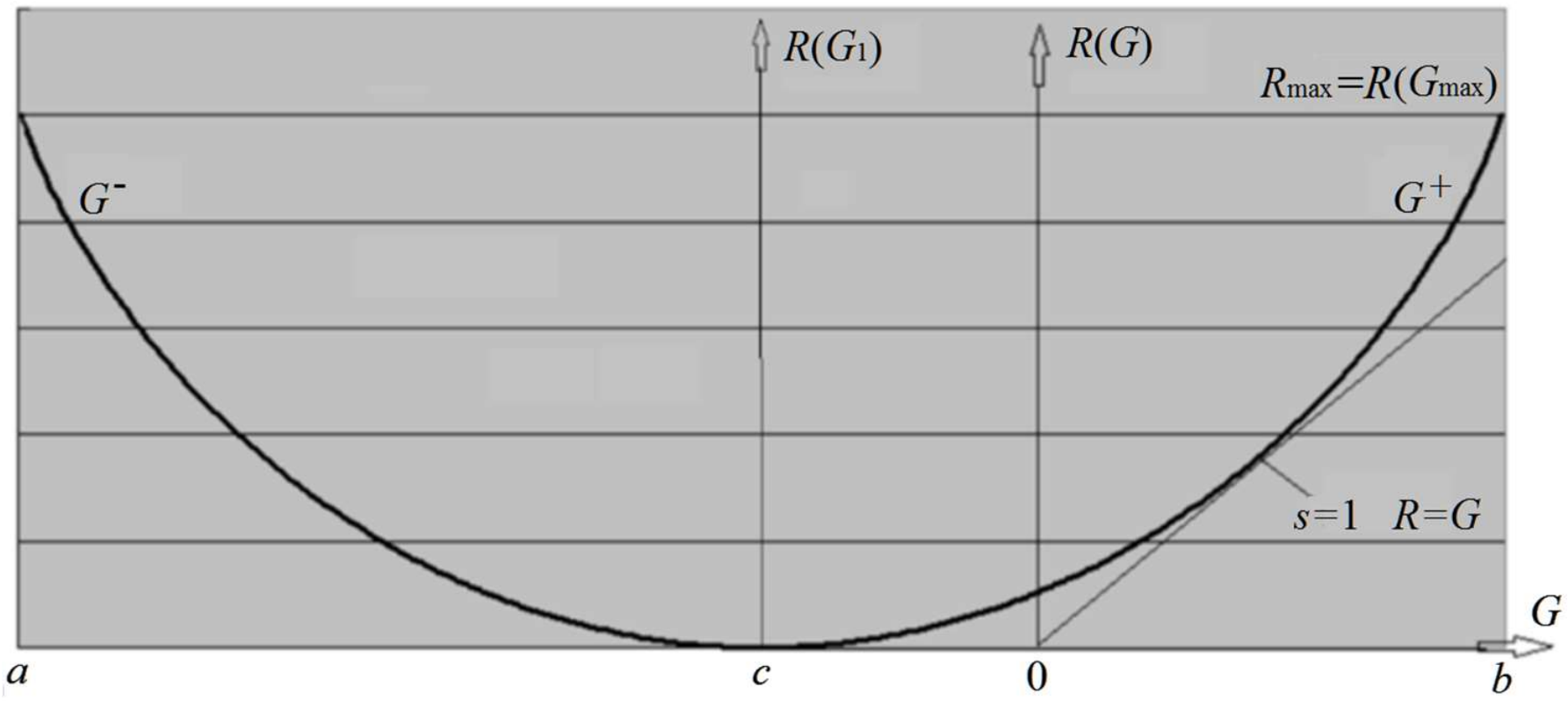

**Figure 5.** The rate-verisimilitude function $R(G)$ of a binary communication. For any $R(G)$ function, there is a point where $R(G)=G$.

In Figure 5, s= d$R$/d$D$. When $s$=1, $R$ is equal to $G$, which means that the semantic channel matches the Shannon channel. $G/R$ indicates the efficiency of the semantic communication. In Section 3.4, we will see that to solve a mixture model is to find a parameter set $\theta$ that maximum $G/R$ so that $G/R$ is close to 1 or $G\approx R$.

When $s\rightarrow\infty$, both $R$ and $G$ reach their maxima $R_{max}$ and $G_{max}$. As $s$ increases, TPFs $P(y_j|x)$, $j$=1,2,…,$n$, will become sharper and the Shannon channel will have less noise. Hence, the $R$ and G will increase. This property of $R(G)$ can be used to prove the convergence of the CM iteration algorithm for the MMI classifications of unseen instances.

The $R$(G) function is different from the $R(D)$ function. For a given $R$, we have the maximum value $G^+$ and the minimum value $G^-$. $G^-$ is negative and means that to bring a certain information loss $|G|$ to enemies, we also need certain objective information $R$. When $R$=0, $G$ is negative, which means that if we listen to someone who randomly predicts, the information that we already have will be reduced.

The $R(G)$ function was developed mainly for image compression according to visual discrimination [25]. Now it can be used for the convergence proofs of MMI classifications and mixture models.

### 2.2. From Traditional Bayes' Prediction to Logical Bayesian Inference

#### 2.2.1. Traditional Bayes' Prediction, Likelihood Inference(LI), and Bayesian Inference (BI)

To understand LBI better, first, we review the traditional Bayes' Prediction (TBP), LI, and BI.

We call the inference with the TPF $P(y_j|x)$ as Traditional Bayes' Prediction (TBP). Using TBP, for given $P(x)$ and $P(y_j|x)$, we can make probability prediction

$$P(x|y_j)=P(x)\,P(y_j|x)/P(y_j). \tag{2.27}$$

When $P(y_j|x)$ is replaced with $kP(y_j|x)$, where $k$ is a constant, $P(x|y_j)$ is the same because

$$\frac{P(x)kP(y_j|x)}{\sum_i P(x_i)kP(y_j|x_i)}=\frac{P(x)P(y_j|x)}{\sum_i P(x_i)P(y_j|x_i)}=P(x|y_j). \tag{2.28}$$

Using this formula, we can easily explain that a truth function that is proportional to a TPF can be used for the same probability prediction.

For given $P(y_j)$, $P(x|y_j)$, and $P(x)$, we can obtain the predictive model, e.g., TPF

$$P(y_j|x)=P(y_j)\,P(x|y_j)/P(x). \tag{2.29}$$

We use the medical test (or the signal detection) as an example to explain how TPFs and the Shannon channel as a predictive model.

**Definition 4**: Let $z$ be an observed feature for an unseen instance (see Figure 1); $Z$ is a random variable taking a value $z\in C=\{z_1, z_2, \ldots\}$. For unseen instance classifications, $x$ denotes a true class or true label.

Assume that we classify every unseen instance with unseen true label $x$ according to its observed feature $z\in C$. That is to provide a classifier $y=f(z)$ to get a label $y$ (see Figure 6).

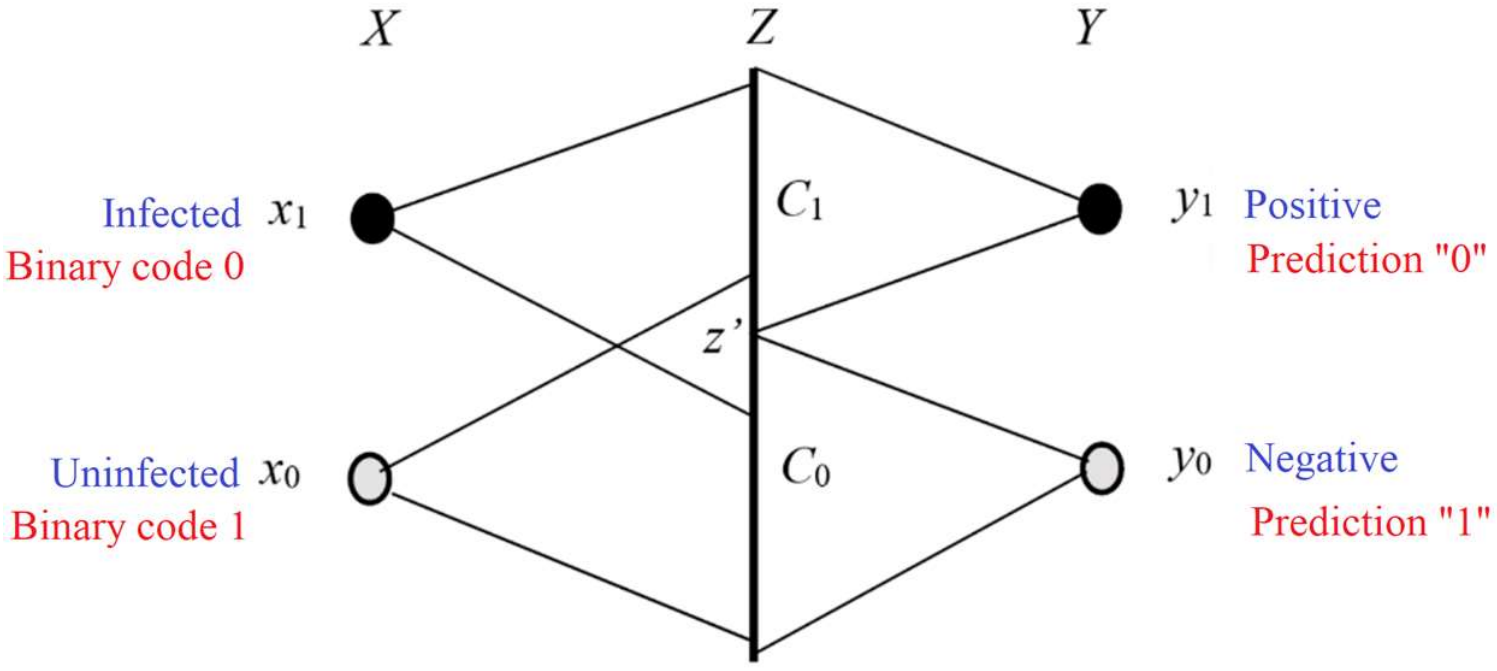

**Figure 5**. Illustrating the medical test and signal detection. We choose $y_j$ according to $z\in C_j$. $\{C_0, C_1\}$ is a partition of $C$.

We use the HIV test to explain that the TPF can be used for probability prediction with different $P(x)$. For an infected testee $x_1$, the conditional probability $P(y_1|x_1)$ of $y_1$=positive is called sensitivity, which means the true positive rate. For an uninfected testee $x_0$, the conditional probability $P(y_0|x_0)$ of $y_0$=negative is called specificity, which means the true negative rate [49]. The sensitivity and specificity ascertain a Shannon channel as shown in Table I.

**Table 1**. The sensitivity and Specificity of a Medical Test ascertain a Shannon's Channel $P(Y|X)$

| | **Infected testee $x_1$** | **Uninfected testee $x_0$** |
|---|---|---|
| Positive $y_1$ | $P(y_1|x_1)$=sensitivity=0.917 | $P(y_1|x_0)$=1-specificity=0.001 |
| Negative $y_0$ | $P(y_0|x_1)$=1-sensitivity=0.083 | $P(y_0|x_0)$=specificity=0.999 |

* Data are from OREQuick HIV tests [50]

**Example 2.2** Calculate $P(x_1|y_1)$ using $P(y_1|x)$ in Table 1 for $P(x_1)$=0.0001, 0.002 (for normal people), and 0.1 (for high-risk crowd).

**Solution:** Using Eq. (2.28), for $P(x_1)$=0.0001, 0.002, and 0.1, we have $P(x_1|y_1)$=0.084, 0.65, and 0.99.

Using Likelihood Inference (LI), it is not easy to solve Example 3.1. Nevertheless, when $x$ is one of many different values, and TPFs are not smooth, we may need LI. If samples are small so that TPFs are not smooth even not continuous, we cannot use TPF to obtain continuous $P(x|y_j)$. That is why we use LI, which uses parameters to construct smooth likelihood functions. Using the MLE, we can train a likelihood function by a sample **X** to obtain the best $\theta_j$:

$$\theta_j{}^* = \arg\max_{\theta_j} P(\mathbf{X}|\theta_j) = \arg\max_{\theta_j} [\sum_i P(x_i|.)\log P(x_i|\theta_j)], \tag{2.30}$$

where $P(x_i|.)$ meas that $y_j$ is unkown. The main defect of LI is that LI cannot make use of prior knowledge, and when $P(x)$ is changed, the optimized likelihood function will be invalid.

To make use of prior knowledge, subjective Bayesians developed Bayesian Inference (BI) [2,3]. They brought the prior distribution $P(\theta)$ of $\theta$ into the Bayes' Theorem to obtain the Bayesian posterior

$$P(\theta|\mathbf{X}) = \frac{P(\theta)P(\mathbf{X}|\theta)}{P_\theta(\mathbf{X})}\ , P_\theta(\mathbf{X}) = \sum_j P(\theta_j)P(\mathbf{X}|\theta_j) \tag{2.31}$$

where $P_\theta(\mathbf{X})$ is the normalized constant related to $\theta$; $P(\theta|\mathbf{X})$ is the posterior distribution of $\theta$ or the Bayesian posterior. Using $P(\theta|\mathbf{X})$, we can derive the Maximum A Posterior estimation:

$$\begin{aligned}\theta_j{}^* &= \arg\max_{\theta_j} P(\theta_j|\mathbf{X}_j) = \arg\max_{\theta} P(\theta_j)P(\mathbf{X}_j|\theta_j) \\ &= \arg\max_{\theta_j}[\sum_i P(x_i|.)\log P(x_i|\theta_j) + \log P(\theta_j)],\end{aligned} \tag{2.32}$$

where $P_\theta(\mathbf{X})$ is neglected.

BI has some advantages, such as

- It is especially suitable to cases where $Y$ is a random variable for a frequency generator, such as a dice.
- As the sample size increases, the distribution $P(\theta|\mathbf{X})$ will gradually shrink to $\theta_j{}^*$ that comes from the MLE.
- BI can make use of prior knowledge better than LI.

However, BI also has some disadvantages:

- The probability prediction according to BI [3] is not compatible with the traditional Bayes' prediction.
- $P(\theta)$ is subjectively selected.
- BI cannot make use of the prior of $x$.

If we try to use BI to solve Example 2.1 and Example 2.2,, we will find that the Bayesian posterior is not as good as TPF $P(y_j|x)$. Therefore, to make use of the prior of $x$, we still want a parameterized TPF $P(\theta_j|x)$.

### 2.2.2. From Fisher's Inverse Probability Function $P(\theta_j|x)$ to Logical Bayesian Inference (LBI)

De Morgan first called TPF $P(y_j|x)$ as "inverse probability" about Laplace's method of probability [2]. The corresponding direct probability is $P(x|y_j)$. Later, Fisher called the likelihood function $P(x|\theta_j)$ as direct probability and the parameterized TPF $P(\theta_j|x)$ as inverse probability [2]. We use $\theta_j$ instead of $\theta$ and $x$ instead of $x_i$ to emphasize $\theta_j$ is a constant and $x$ is a variable, and hence, $P(\theta_j|x)$ should be a function. In the following, we call $P(\theta_j|x)$ as the Inverse Probability Function (IPF). According to Bayes' theorem, there is

$$P(\theta_j|x) = P(\theta_j)\ P(x|\theta_j)/P(x), \tag{2.33}$$

$$P(x|\theta_j)=P(x_i)\ P(\theta_j|x)/P(\theta_j). \tag{2.34}$$

IPF $P(\theta_j|x)$ can make use of the prior knowledge $P(x)$ well. When $P(x)$ becomes $P'(x)$, we can still obtain $P'(x|\theta_j)$ from $P'(x)$ and $P(\theta_j|x)$.

When $n$=2, we can easily construct $P(\theta_j|x)$, $j$=1,2, with parameters. For instance, we can use a pair of Logistic (Sigmoid) functions as the IPFs. Unfortunately, when $n$>2, it is hard to construct $P(\theta_j|x)$, $j$=1,2,…,$n$, because there is normalization limitation $\sum_j P(\theta_j|x)=1$
for every $x$. That is why a multi-class or Multilabel classification is often converted into several binary classifications [51,52].

$P(\theta_j|x)$ and $P(y_j|x)$ as predictive models also have a serious disadvantage. In many cases, we can only know $P(x)$ and $P(x|y_j)$ without knowing $P(\theta_j)$ or $P(y_j)$ so that we cannot obtain $P(y_j|x)$ or $P(\theta_j|x)$. Nevertheless, we can obtain truth function $T(\theta_j|x)$ in these cases. For LBI, there is not the normalization limitation, and hence it is easy to construct a group of truth functions and train them with $P(x)$ and $P(x|y_j)$, $j$=1,2,…,$n$, without $P(y_j)$ or $P(\theta_j)$. That is an important reason why we need LBI.

When a sample $\mathbf{X}_j$ is huge, we can directly obtain $T^*(\theta_j|x)$ from Eq. (2.20). For a size-limited sample, we can use the generalized KL to obtain

$$\begin{aligned} T^*(\theta_j|x) &= \underset{T(\theta_j|x)}{\arg\max}\, I(X;\theta_j) = \underset{T(\theta_j|x)}{\arg\max} \sum_i P(x_i|y_j)\log\frac{T(\theta_j|x_i)}{T(\theta_j)} \\ &= \underset{T(\theta_j|x)}{\arg\max} \sum_i P(x_i|y_j)\log\frac{P(x_i|\theta_j)}{P(x_i)}. \end{aligned} \tag{2.35}$$

This formula is the main formulas used for LBI. LBI provides the Maximum Semantic Information Estimation (MSIE) :

$$\begin{aligned} \theta_j^* &= \underset{\theta_j}{\arg\max}\, I(X;\theta_j) = \underset{\theta_j}{\arg\max}\Big[\sum_i P(x_i|.)\log[P(x_i|\theta_j)/P(x_i)]\Big] \\ &= \underset{\theta_j}{\arg\max}\Big[\sum_i P(x_i|.)\log T(\theta_j|x_i) - \log T(\theta_j)\Big], \end{aligned} \tag{2.36}$$

which is compatible with MLE. If samples are big enough, MSIE, MLE, and MAP are equivalent.

We suggest using the truth function as the predictive model or the inference tool for LBI in some cases because the truth function has the following advantages:

- We can use an optimized truth function $T^*(\theta_j|x)$ to make probability prediction for different $P(x)$ as well as we use $P(y_j|x)$ or $P(\theta_j|x)$.
- We can train a truth function with parameters by a sample with small size as well as we train a likelihood function.
- The truth function can indicate the semantic meaning of a hypothesis or the denotation of a label.
- It is also the membership function, which is suitable for classification.
- To train a truth function $T(\theta_j|x)$, we only need $P(x)$ and $P(x|y_j)$, without needing $P(y_j)$ or $P(\theta_j)$.
- Letting $T(\theta_j|x)\propto P(y_j|x)$, we can set a bridge between statistics and logic.

The CM algorithms can further reveal these advantages.

## 3. Methods II: The Channel Matching (CM) Algorithms

### 3.1. CM1: To Resolve the Multilabel-Learning-for-New-*P*(*x*) problem

#### 3.1.1. For the Optimization of Truth Functions or Membership Functions

We use CM1 to denote the basic matching algorithm in which the semantic channel matches the Shannon channel, where membership functions or truth functions are used as learning functions.

Assume that $x$ is an age; $y_j$ is a label "Youth"; $\theta_j$ is a fuzzy set $\{x|x$ is a youth$\}$. From population statistics, we can obtain population age distribution $P(x)$ and posterior distribution $P(x|y_j)$.

If the sample is huge and hence distributions $P(x)$ and $P(x|y_j)$ are smooth, we can directly use Eq. (2.20) to obtain the optimized membership function $T^*(\theta_j|x)$ without parameters. If $P(x)$ and $P(x|y_j)$ are not smooth, we can use Eq. (2.35) to obtain $T^*(\theta_j|x)$ with parameters. Without needing $P(y_j)$, CM1 for every label's learning is independent.

If the given sampling distribution is transition probability function $P(y_j|x)$, we may assume $P(x)$ is flat. Then Eq. (2.35) becomes

$$T^*(\theta_j|x)=\underset{T(\theta_j|x)}{\arg\max}\sum_i \frac{P(y_j|x_i)}{\sum_k P(y_j|x_k)}\log\frac{T(\theta_j|x_i)}{\sum_k T(\theta_j|x_k)}. \tag{3.1}$$

If $P(y_j|x)$ is smooth, we can use Eq. (2.21) to obtain $T^*(\theta_j|x)$ without parameters. For multilabel learning, we can directly obtain a group of truth functions from a Shannon channel $P(Y|X)$ or a sample with distribution $P(x,y)$. However, using the Binary Relevance, we have to prepare several samples for several Logistic functions.

When $P(x)$ is changed, $T^*(\theta_j|x)$ is still useful to makes semantic Bayes' predictions.

3.1.2. For the Confirmation Measure of Major PremisesWe use "degree of confirmation" to denote "degree of belief" supported by evidence or samples.

Bayesians use "degree of belief" to explain the subjective probability of a hypothesis. This degree of belief is between 0 and 1. However, the researchers of induction use "degree of belief" to evaluate if-then statements or major premises. This degree of belief should be between -1 and 1. In this paper, we use "degree of belief" between -1 and 1 for the subjective evaluation of if-then statements. We know that the correlation coefficient between two random variables is also between -1 and 1. The difference is that if-then statements are asymmetry; there are more than one major premises and degrees of belief between instance $X$ and hypothesis $Y$.

Now we use the medical test as an example to explain how to use truth functions to replace TPFs or how to use the semantic channel to replace the Shannon channel for probability predictions.

From the Shannon channel in Table 1, we can derive the semantic channel as shown in Table 2. Assume $T(y_1|x_1)=T(y_0|x_0)=1$ and $T(y_1|x_0)=T(y_0|x_1)=0$ for non-fuzzy hypotheses. Two truth functions for corresponding fuzzy hypotheses are

$$T(\theta_1|X)= b_1' +b_1T(y_1|x), \tag{3.2}$$

$$T(\theta_0|X)= b_0' +b_0T(y_1|x), \tag{3.3.3}$$

where $b_1=b(y_1\rightarrow x_1)$ means the degree of belief of major premise $MP_1= y_1\rightarrow x_1$ ="if $Y=y_1$ then $X=x_1$",, and $b_1'=1-|b_1|$ means the degree of disbelief of $MP_1$ and the ratio of a tautology in $y_1$. Likewise, $b_0$ $b(y_0\rightarrow x_0)$ and $b_0'=1-|b_0|$.

**Table 2**. Two degrees of disbelief of a medical test form a semantic channel $T(\theta|x)$

| $Y$ | Infected $x_1$ | Uninfected $x_0$ |
|---|---|---|
| Positive $y_1$ | $T(\theta_1\|x_1)=1$ | $T(\theta_1\|x_0)=b_1'$ |
| Negative $y_0$ | $T(\theta_0\|x_1)=b_0'$ | $T(\theta_0\|x_0)=1$ |

According Eqs. (220) and (2.21), two optimized degrees of disbelief are

$$b_1'^*= P(y_1|x_0)/ P(y_1|x_1)=[P(x_0|y_1)/P(x_0)]/[P(x_1|y_1)/P(x_1)], \tag{3.4}$$

$$b_0'^*= P(y_0|x_1)/P(y_0|x_0)= [P(x_1|y_0)/P(x_1)]/[P(x_0|y_0)/P(x_0)]. \tag{3.5}$$

For given $y_1$, we can use $b_1'^*$ and different $P(x)$ to make the semantic Bayesian prediction:

$$P(x_1|\theta_1)=P(x_1)/[P(x_1)+b_1'^*P(x_0)], \tag{3.6}$$

$$P(x_0|\theta_1)= b_1'^*P(x_0) /[P(x_1)+b_1'^*P(x_0)]. \tag{3.7}$$

This prediction is equivalent to the traditional Bayes' prediction with TPF $P(y_j|x)$. Even if we only know $P(x|y_1)$ and $P(x)$ without knowing $P(y_1)$, we can still make the probability prediction. It is easy to verify that, using Eq. (3.6) to solve Example 2.2, the results are the same as those from the traditional Bayes' prediction.

If we try to use LI or BI to solve Example 3.1, we will find it is not easy for the model to fit different $P(x)$.

In comparison with the Shannon channel in Table 1, the semantic channel in Table 2 is easier to understand and remember. To remember $P(y_j|x)$, we need to remember two numbers; whereas, to remember $T^*(\theta_1|x)$, we only need to remember one number $b'^*$.

In reference [41], the author provides two formulas for positive and negative degrees of confirmation respectively. Now we can merge the two formulas into a new formula:

$$\begin{aligned} b_1^* = b^*(y_1 \rightarrow x_1) &= \frac{P(y_1|x_1) - P(y_1|x_0)}{\max(P(y_1|x_1), P(y_1|x_0))} = \frac{\text{sensitivity-(1-specificity)}}{\max(\text{sensitivity,1-specificity})} \\ &= \frac{\text{True_positive_rate} - \text{False_positive_rate}}{\max(\text{True_positive_rate, False_positive_rate})} = \frac{CL_1 - CL_1'}{\max(CL_1, CL_1')}, \end{aligned} \tag{3.8}$$

where $CL_1 = P(y_1|x_1)/[P(y_1|x_0)+P(y_1|x_1)]$ is the confidence level of $MP_1$ and $CL_1' = 1 - CL_1$. As $CL_1$ changes from 0 to 1, $b_1^*$ changes from -1 to 1 as shown in Figure 7.

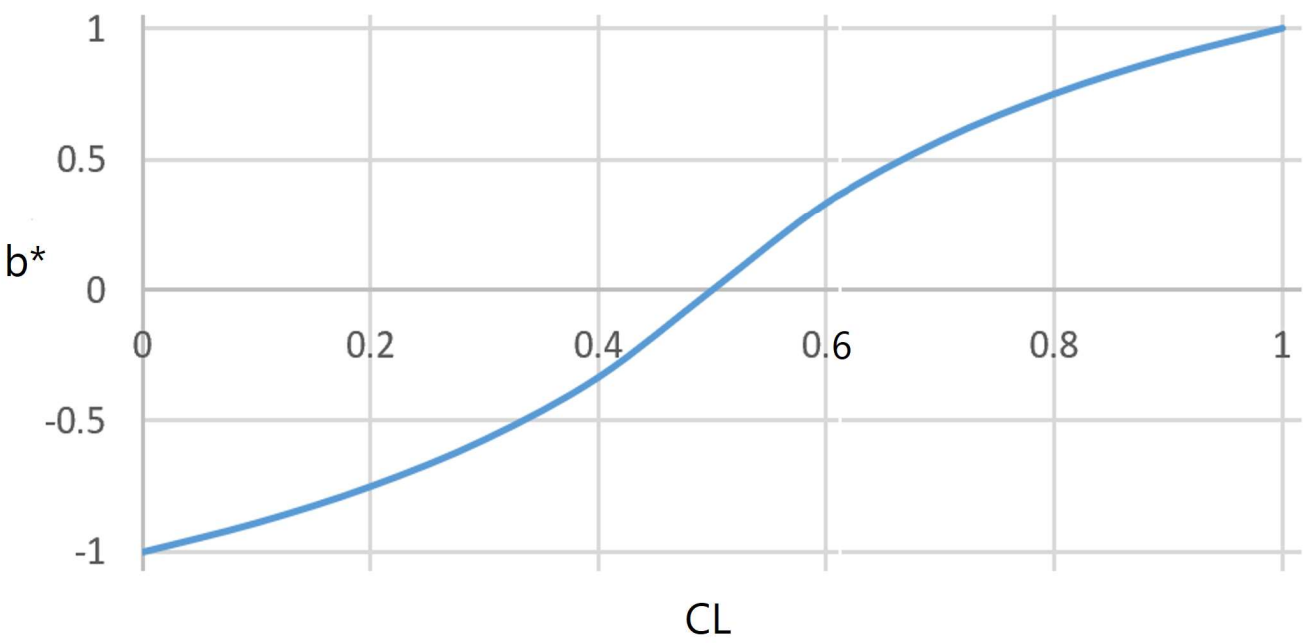

**Figure 7.** Relationship between degree of conformation $b^*$ and confidence level $CL$. As $CL$ changes from 0 to 1, $b^*$ changes from -1 to 1.

### 3.1.3. For Rectifying the Parameters of a GPS Device

If we do not know the real parameters of a GPS device or suspect the parameters claimed by the producer, we can assume

$$T(\theta_j|x) = \exp[-(x - x_j - \Delta x))^2/(2\sigma^2)], \tag{3.9}$$

wher $x$ is a two-dimensional vector. Then we can use a sample to find parameters $\Delta x$, which is the systematic deviation, and $\sigma$. We may obtain the sample by driving a car with the GPS device on a big square and recording the relative positions $x' = x - x_j$. From many relative deviations, we can obtain sampling distribution $P(x'|y_j)$. Since driving on a big square, $P(x)$ should be flat. Then we can use the generalized KL information formula to obtain the optimized parameters $\Delta x^*$ and $\sigma^*$. We can use $\Delta x^*$ to adjust $y_j$ by replacing $y_j$ with $y_k = y_j + \Delta x^*$.

Assume that the GPS device is often faulty, we can also use

$$T(\theta_j|x) = b\exp[-(x - x_j - \Delta x)^2/(2\sigma^2)] + 1 - b \tag{3.10}$$

as the learning function to obtain the confirmation measure $b^*$ of the GPS device.

If one tries to use the inverse likelihood function $P(\theta_j|x)$ or the Bayesian posterior $P(\theta|\mathbf{X})$ for the same task and probability prediction (see Figure 3), he will find it is very difficult to do because we only have prior knowledge $P(x)$ from a GPS map, without the prior knowledge $P(y)$ or $P(\theta)$.

### MultilabelMultilabel3.2. CM2: The Semantic Channel and the Shannon Channel Mutually Match for Multilabel Classifications

We use CM2 to denote two steps:

- **Matching I:** Let the semantic channel match the Shannon channel or use CM1 for multilabel learning;
- **Matching II:** Let the Shannon channel match the semantic channel by using the Maximum Semantic Information (MSI) classifier.

Both steps use the MMI or ML criterion.

MultilabelMultilabelFor Multilabel learning, we may train every label by Eq. (2.35) or Eq. (3.1). We may also learn from the popular method [52] to use both $P(x|y_j)$ and $P(x|y_j')$, where $y_j'$ is the negation of $y_j$, to train $T(\theta_j|x)$ by

$$\begin{aligned} T^*(\theta_j|x) &= \arg\max_{T(\theta_j|X)} [I(X;\theta_j) + I(X;\theta_j^c)] \\ &= \arg\max_{T(\theta_j|x)} \sum_i [P(x_i|y_j)\log\frac{T(\theta_j|x_i)}{T(\theta_j)} + P(x_i|y_j{}')\log\frac{1-T(\theta_j|x_i)}{1-T(\theta_j)}], \end{aligned} \tag{3.11}$$

where $\theta_j^c$ is the complementary set of $\theta_j$. This $T^*(\theta_j|x)$ may be a Logistic function,which will cover a larger area of $U$ in comparison with $T^*(\theta_j|x)$ from Eq. (2.35) or Eq. (3.1).

If there are examples with one instance and several labels, or with several instances and one label, we can split such an example into several single-instance and single-label examples as well as the popular method does [51]. Then we can obtain the Shannon channel $P(Y|X)$ for multilabel learning.

MultilabelMultilabelMultilabelMultilabelFor seen instance classifications, $x$ is given.

For Matching II, the MSI classifier is

$$y_j^* = h(x) = \arg\max_{y_j} \log I(x;\theta_j) = \arg\max_{y_j} \log[T(\theta_j|x)/T(\theta_j)]\,. \tag{3.12}$$

Using $T(\theta_j)$, we can overcome the class-imbalance problem [50] and reduce the rate of failing to report smaller probability events. If $T(\theta_j|X) \in \{0,1\}$, the semantic information measure becomes Carnap and Bar-Hillel's semantic information measure, and the classifier becomes the minimum logical probability classifier:

$$y_j^* = h(x) = \arg\max_{y_j \text{ with } T(A_j|x)=1} \log[1/T(A_j)] = \arg\min_{y_j \text{ with } T(A_j|x)=1} \log T(\theta_j)\,. \tag{3.13}$$

This criterion encourages us to select a compound label with smaller denotation.

For unseen instance classifications or uncertain $x$, we only know $P(x|z)$. The MSI classifier becomes

$$y_j^* = f(z) = \arg\max_{y_j} \sum_i P(x_i|z)\log\frac{T(\theta_j|x_i)}{T(\theta_j)}. \tag{3.14}$$

To simplify Multilabel learning, we may train fewer atomic labels and use them and the fuzzy logic that is compatible with Boolean algebra [22] Multilabelto produce the membership function of a compound label for the multilabel classifications [44].

In the popular method for multilabel classifications with the Bayes' classifier or MPP criterion, for different $x$, the classifier compares two IPFs $P(\theta_j|x)$ and $P(\theta_k|x)$, such as two Logistic functions, to select a label with greater IPF. This method is not compatible with the information criterion or the likelihood criterion.

Multilabel

### 3.3. CM3: the CM Iteration Algorithm for the MMI classifications of Unseen instances

We use CM3 to denote the CM iteration algorithm, which repeats two steps: Matching I and Matching II. CM2 is not an iterative algorithm, nevertheless, CM3 is. This algorithm is used for the MMI classifications, for which the popular method is the Gradient Decent.

We use the medical test as shown in Figure 6 as an example to explain the problem with the MMI classifications of unseen instances.

We need to optimize $z'$ for the MMI. The problem is that, without the classifier $f(z)$, we cannot express mutual information $I(X; Y)$, whereas, without the expression of mutual information, we cannot optimize the classifier $f(z)$. This problem is also met by the MLE for uncertain Shannon's channels. To resolve this problem, researchers generally use parameters to construct construct partition boundaries and then use the Gradient Descent or the Newton method to search the best parameters for MMI. The CM iteration algorithm for the MMI classifications is different. It uses numerical values to express boundaries and information gain functions, e.g., reward functions. It repeats updating information gain functions and boundaries to achieve MMI.

Let $C_j$ be a subset of $C$ and $y_j$=$f(z|z \in C_j)$; hence $S$={$C_1$, $C_2$, …} is a partition of $C$. Our aim is, for given $P(x, z)$ from **D**, to find the optimized $S$, which is

$$S^* = \arg\max_S I(X;\theta|S) = \arg\max_S \sum_j \sum_i P(C_j)P(x_i \mid C_j)\log\frac{T(\theta_j \mid x_i)}{T(\theta_j)}. \quad (3.15)$$

**Matching I**: Let the semantic channel match the Shannon channel and set reward functions..

First, we obtain the Shannon channel for given $S$:

$$P(y_j \mid x) = \sum_{z_k \in C_j} P(z_k \mid x),\ j = 1, 2, ..., n. \quad (3.16)$$

From this Shannon's channel, we can obtain the semantic channel $T(\theta|X)$ and the semantic information $I(x_i; \theta_j)$. Then, for given $z$, we have conditional information or reward functions:

$$I(X_i;\theta_j|z) = \sum_i P(x_i \mid z) I(x_i;\theta_j)\ ,\quad j=0,1,\ldots,n, \quad (3.17)$$

which are some curved surfaces over a two-dimensional feature space when $U$ is two-dimensional. We may directly let $I(x_i; \theta_j)$=$I(x_i; y_j)$=$\log[P(y_j|x)/P(y_j)]$. However, with the notion of the semantic channel, we can understand this algorithm and prove its convergence better.

**Matching II:** Let the Shannon channel match the semantic channel by the classifier

$$y_j^* = f(z) = \arg\max_{y_j} I(X_i;\theta_j|z),\ j=0,1,\ldots,n. \quad (3.18)$$

Repeat Matching I and II until $S$ does not change. The convergent $S$ is $S^*$ we seek.

Matching II for the optimization of the Shannon channel can reduce noise. We can understand two matching steps in this way: Matching I is for the reward function; Matching II is for Bayes' decision.

For a given source $P(X)$, a semantic channel ascertains an $R(G)$ function. An improved $R(G)$ function has a higher matching point where $R(G)$=$G$. The CM algorithm is to find the matching point that is also the point with $R_{max}$ and $G_{max}$ (see Figure 8).

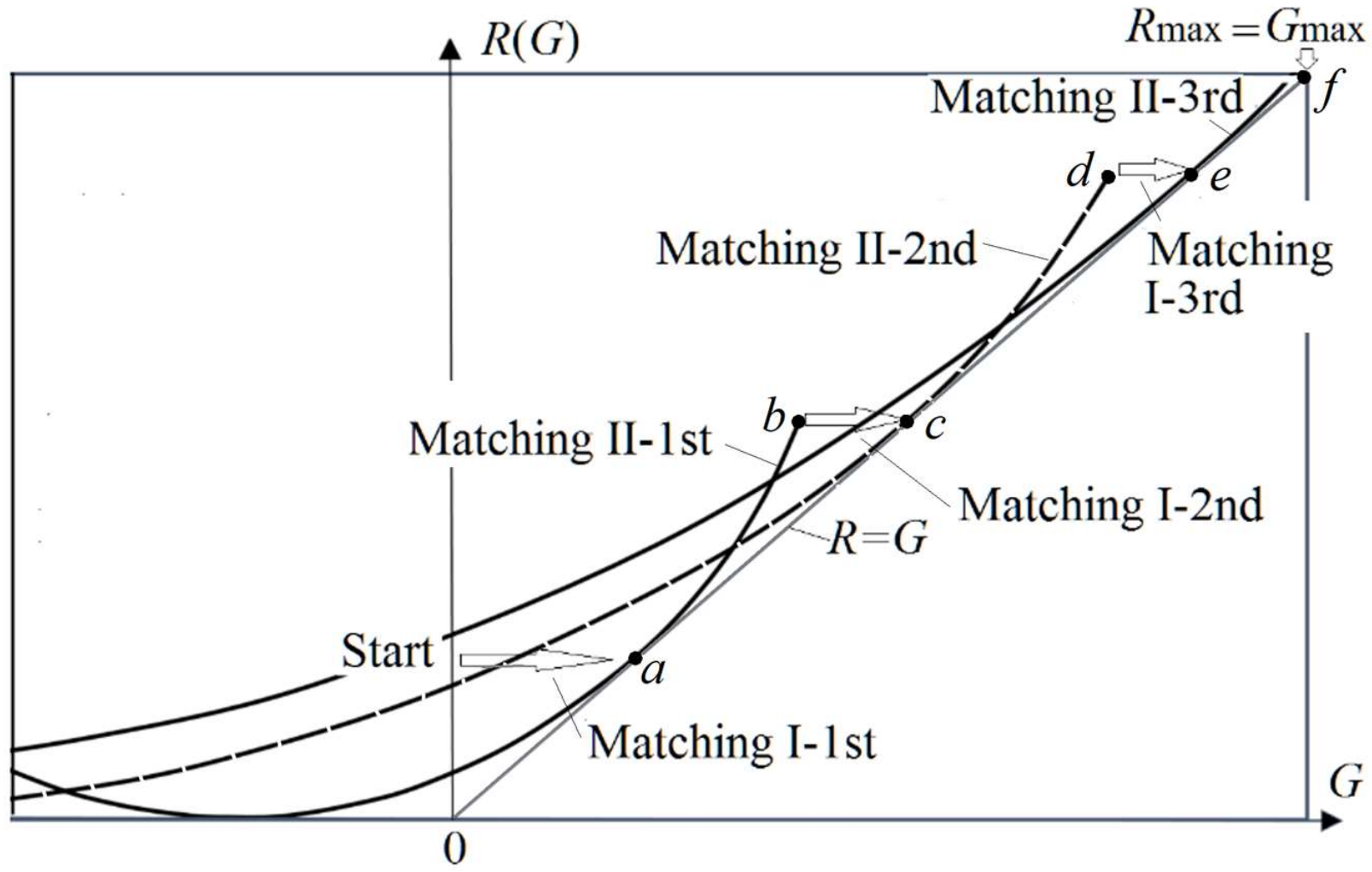

**Figure 8.** Illustrating the iterative convergence of the MMI classifications of unseen instances. In the iterative process, ($G$,$R$) moves from the start point to $a$,$b$,$c$,$d$,$e$,…,$f$ gradually.

.

We can prove that the iteration can converge. In the iterative process of the CM iteration algorithm, coordinate ($G$, $R$) changes as follows:

$$\text{Start} \xrightarrow{\text{Matching I-1st}} a \xrightarrow{\text{Matching II-1st}} b \xrightarrow{\text{Matching I-2nd}} c \xrightarrow{\text{Matching II-2nd}} d$$

$$\xrightarrow{\text{Matching I-3rd}} e \xrightarrow{\text{Matching II-3rd}} \ldots f$$

The process continues until Matching II cannot improve $R$ and $G$.

The coordinate ($G$, $R$) can converge to ($G_{\max}$, $R_{\max}$) because every Matching I increases $G$ and every Matching II increases $G$ and $R$, and the maxima of $G$ and $R$ are finite.

Matching II can always find any best partition for given $I(X; y_j|z)$, $j$=1,2,… because it checks every $z$ to see which of $I(X; y_j|z)$, $j$=1,2,… is maximum.We can understand the CM algorithm in the following way. An $R(G)$ function is like a ladder, and coordinate ($G$, $R$) is like a climber. In Matching I, ($G$, $R$) creates a ladder and moves on it. In Matching II, it climbs up to the top of the ladder. Then it creates a new ladder … until it reaches ($G_{\max}$, $R_{\max}$).

### 3.4. CM4: the CM-EM Algorithm for Mixture Models

We use CM4 to denote the CM-EM Algorithm: An Improved EM Algorithm for mixture models.

In CM3, Matching II is for maximum $R$, whereas, in CM4, Matching II is for maximum information efficiency $G/R$ or minimum $R$-$G$.

CM4 is based on a totally different convergence theory of mixture models. The popular convergence theory of the EM algorithm explains that we can maximize incomplete data log-likelihood $L_X(\theta)$ by maximizing complete data log-likelihood $Q$, whereas the convergence theory basing the CM-EM algorithm explains that we can maximize $L_X(\theta)$ by maximizing information efficiency $G/R$.

The EM algorithm [53] for mixture models often results in slow or invalid convergence [54,55]. We can improve the EM algorithm by letting the semantic channel and the Shannon channel mutually match. The different is that Matching II is for the minimum of Shannon's mutual information $R$.

If a probability distribution $P_\theta(x)$ comes from the mixture of $n$ likelihood functions, e. g.,

$$P_\theta(x) = \sum_{j=1}^{n} P(y_j)P(x \mid \theta_j), \tag{3.19}$$

then we call $P_\theta(x)$ a mixture model. If every predictive model $P(x|\theta_j)$ is a Gaussian function, then $P_\theta(x)$ is a Gaussian mixture. In the following, we use $n$=2 to discuss the algorithms for mixture models.

Assume $P(x)$ comes from the mixture of two true model $P(x|\theta_1{}^*)$ and $P(x|\theta_2{}^*)$ with ratios $P^*(y_1)$ and $P^*(y_2)=1-P^*(y_1)$. That is

$$P(x)=P^*(y_1)P(x|\theta_1{}^*)+P^*(y_2)P(x|\theta_2{}^*). \tag{3.20}$$

We only know $P(x)$ and $n$=2. We can use guessed parameters and and the mixture ratios to obtain

$$P_\theta(x)=P(y_1)P(x|\theta_1)+P(y_2)P(x|\theta_2). \tag{3.21}$$

Then we have log-likelihood

$$L_X(\theta) = N\sum_i P(x_i)\log P_\theta(x_i) = -NH_\theta(X), \tag{3.22}$$

and the relative entropy or the KL divergence:

$$H(P \,\|\, P_\theta) = \sum_i P(x_i)\log\frac{P(x_i)}{P_\theta(x_i)} = H_\theta(X) - H(X). \tag{3.23}$$

If two distributions P(x) and $P_\theta$(x) are close to each other so that the relative entropy is close to 0, such as less than 0.001 bit, then we may say that our guess is right. Therefore, our task is to change $\theta$ and $P(y)$ to maximize likelihood $L_X(\theta)=\log P(\mathbf{X}|\theta)$ or to minimize relative entropy $H(P||P_\theta)$.

The main formula of the EM algorithm for mixture models can be described as follows:

(3.24)

$$\begin{aligned} Q &= N\sum_i\sum_j P(x_i)P(y_i \mid x_i)\log P(x_i, y_j \mid \theta) \\ &= L_X(\theta) + N\sum_i\sum_j P(x_i)P(y_j \mid x_i)\log P(y_j \mid x_i), \end{aligned} \tag{3.24}$$

where $Q= -NH(X, Y|\theta)$ is called as complete data log-likelihood. $P(y_j|x)$ is from Eq. (3.26). There is

$$L_X(\theta)=Q-H, \tag{3.25}$$

where $H=-NH(Y|X,\theta)$ is a Shannon conditional entropy.

Steps in the EM algorithm are:

**E-step:** Write the conditional probability functions (e. g., the Shannon channel):

$$\begin{aligned} &P(y_j \mid x) = P(y_j)P(x \mid \theta_j) / P_\theta(x), \\ &P_\theta(x) = \sum_j P(y_j)P(x \mid \theta_j). \end{aligned} \tag{3.26}$$

**M-step**: Improve $P(y)$ and $\theta$ to maximize $Q$. If $Q$ cannot be improved further, then end the iteration process; otherwise, go to the E-step.

Neal and Hinton [56] proposed an improved EM algorithm, the Maximization-Maximization (MM) algorithm, in which $Q$ is replaced with $F=Q+H(Y)$. Both steps maximize $F$.

Almost all the EM algorithm researchers believe that $Q$ and $\log L_X(\theta)$ are positively correlated and the E-step does not decrease $Q$; nevertheless, it is not true. The author found that $Q$ may decrease in some E-steps; $Q$ and $F$ should decrease in some cases [42].

Using the CM algorithm to improve the EM algorithm, we can have an algorithm, the CM-EM algorithm, for better convergence.

The CM-EM algorithm includes three steps:

**E1-step:** Construct the Shannon channel. This step is the same as the E-step of the EM algorithm.
**E2-step:** Repeat the following three equations until $P^{+1}(y)$ converges to $P(y)$.

$$\begin{aligned}&P^{+1}(y_j)=\sum_i P(x_i)P(y_j\,|\,x_i)=\sum_i P(x_i)P(y_j\,|\,x_i),\ j=1,2,...;\\&P(y_j)=P^{+1}(y_j);\\&P(y_j\,|\,x_i)=P(x_i\,|\,\theta_j)P(y_j)/\sum_k P(y_k)P(x_i\,|\,\theta_k),\ i=1,2,...;\ j=1,2,...\end{aligned}\tag{3.27}$$

If $H(P||P_\theta)$ is less than a small value, then end the iteration.

**MG-step:** Optimize the parameters $\theta_j^{+1}$ of the likelihood function in log(.) to maximize $G$:

$$G=I(X;\theta)=\sum_i\sum_j P(x_i)\frac{P(x_i\,|\,\theta_j)}{P_\theta(x_i)}P(y_j)\log\frac{P(x_i\,|\,\theta_j^{+1})}{P(x_i)}.\tag{3.28}$$

Then go to E1 -step.

Since $G$ reaches the maximum when $P(x|\theta_j^{+1})/P(x)=P(x|\theta_j)/P_\theta(x)$, the new likelihood function is

$$P(x|\theta_j^{+1})=P(x)P(x|\theta_j)/P_\theta(x).\tag{3.29}$$

Without E2-step, $P(x|\theta_j^{+1})$ above is not normalized in general [61]. For Gaussian mixtures, we can easily obtain new parameters:

$$\begin{aligned}&\mu_j^{+1}=\sum_i P(x_i)P(x_i\,|\,\theta_j)/P_\theta(x_i),\ j=1,\ 2,...,n;\\&\sigma_j^{+1}=\{\sum_i P(x_i)[P(x_i\,|\,\theta_j^{+1})-\mu_j^{+1}]^2\}^{0.5},\ j=1,\ 2...,n.\end{aligned}\tag{3.30}$$

If the likelihood functions are not Gaussian distributions, we can find optimized parameters by searching the parameter space, such as using the Gradient Descent.

To prove the convergence of the CM-EM algorithm, we may make use of the properties of the $R(G)$ function:

- $R(G)$ function is concave and $R(G)$-$G$ has the exclusive minimum 0 as $R(G)=G$ [25];
- $R(G)$-$G$ is close to relative entropy $H(P||P_\theta)$.

After E1 -step, Shannon's mutual information $I(X;\,Y)$ becomes

$$R=\sum_i\sum_j P(x_i)\frac{P(x_i\,|\,\theta_j)}{P_\theta(x_i)}P(y_j)\log\frac{P(y_j\,|\,x_i)}{P^{+1}(y_j)}.\tag{3.31}$$

We define

$$R''=\sum_i\sum_j P(x_i)\frac{P(x_i\,|\,\theta_j)}{P_\theta(x_i)}P(y_j)\log\frac{P(x_i\,|\,\theta_j)}{P_\theta(x_i)}.\tag{3.32}$$

It is easy to prove that $R''$-$G$=$H(P||P_\theta)$. Hence

$$H(P\,||\,P_\theta)=R''-G=R-G+H(Y\,||\,Y^{+1}),\tag{3.33}$$

where

$$H(Y^{+1}||Y)=\sum_j P^{+1}(y_j)\log[P^{+1}(y_j)/P(y_j)].\tag{3.34}$$

Proving that $P_\theta(X)$ converges to $P(X)$ is equivalent to proving that $H(P||P_\theta)$ converges to 0. Since E2-step makes $R$=$R''$ and $H(Y^{+1}||Y)$=0, we only need to prove that every step minimizes $R$-$G$. It is evident that the MG-step minimizes $R$-$G$ because this step maximizes $G$ without changing $R$. The remaining problem is how to prove that E1-step and E2-step minimize $R$-$G$. Learning from the variational method and the iterative method that Shannon [30] and others [48] used for analyzing the rate-distortion function $R(D)$, we can optimize $P(y|x)$ and $P(y)$ respectively to minimize $R\text{-}G=I(X; Y)\text{-}I(X; \theta)$. Since $P(Y|X)$ and $P(Y)$ are interdependent, we can only fix one to optimize another. E2-step is for this purpose. For the detailed convergence proof, see [57].

## 4. Results

### 4.1. The Results of CM2 for Multilabel Learning and Classification

We use a prior distribution $P(x)$ and a posterior distribution $P(x|y_j)$ to optimize a truth function to obtain $T^*(\theta_j|x)$ as shown in Figure 9.

For $P(x)$ and $P(x|y_j)$, first we use a Gaussian random number generator to produce two samples $S_1$ and $S_2$. Both sample sizes are 100000. The data with distribution $P(x)$ is a part of $S_1$. There is

$$P(x) \approx \begin{cases} k\exp[-(x/40)^2], & 0 \le x \le 86; \\ 0, & \text{otherwise.} \end{cases}$$

The $k$ is a normalizing constant. $S_2$ has distribution $P_2(x)$. $P(x|y_j)$ is produced from $P_2(x)$ and the following truth function:

$$T(\theta_2|x) = \begin{cases} \exp[-(x-18)^2/(2*3^2)], \ x < 18; \\ 1, \ 18 \le 1 \le 25; \\ \exp[-(x-25)^2/(2*4^2)], \ x > 25. \end{cases}$$

That means $P(x|y_j)=P_2(x|\theta_2)=P_2(x)T(\theta_2|x)/T(\theta_2)$.

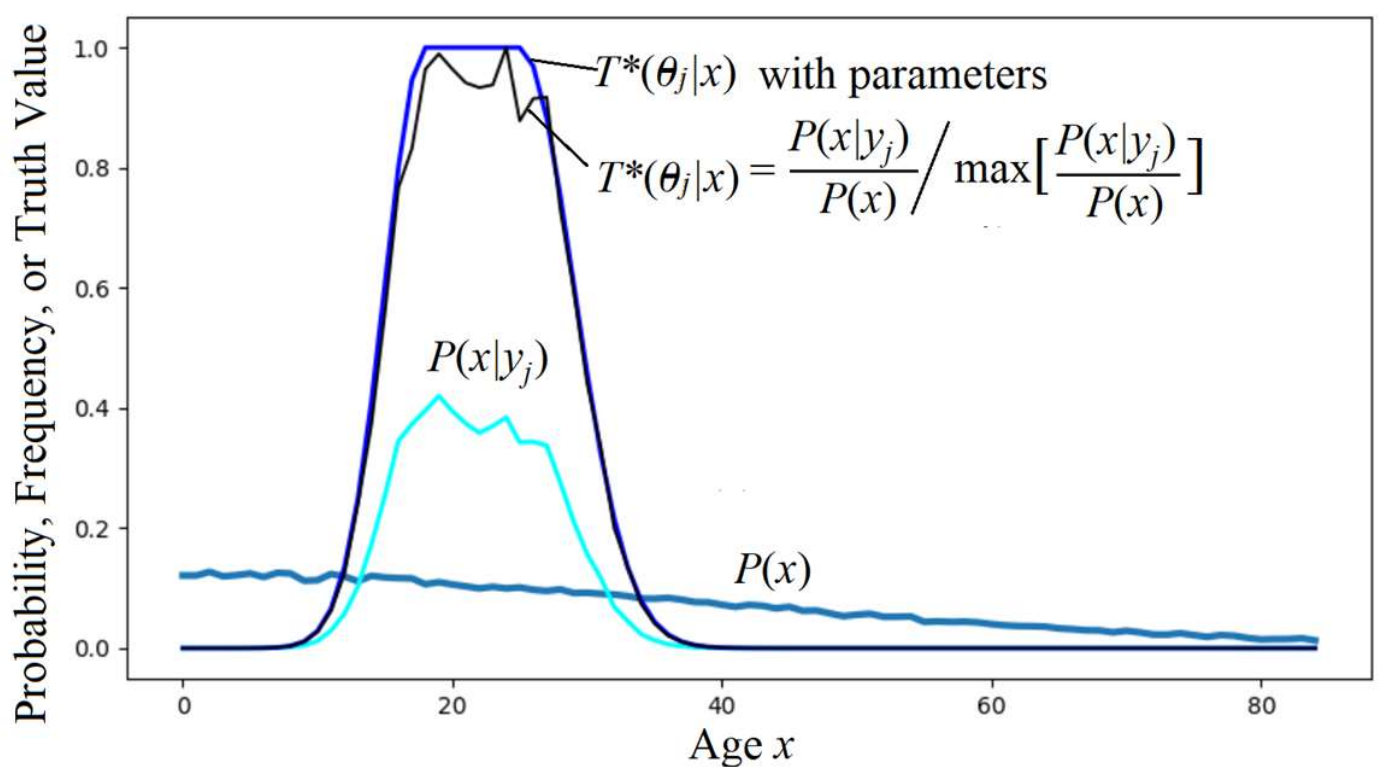

**Figure 9.** Using prior and posterior distributions $P(x)$ and $P(x|y_j)$ to obtain the optimized truth function $T^*(\theta_j|x)$. Figures 9-15 are produced by Python 3.6 files that can be find through Appendix B.

Then we try to obtain $T^*(\theta_j|x)$ from $P(x)$ and $P(x|y_j)$. If we directly use formula Eq. (2.20), $T^*(\theta_j|x)$ will be unsmooth. We can set a truth function with parameters

$$T(\theta_j|x) = \begin{cases} \exp[-(x-\mu_{j1})^2/(2\sigma_{j1}{}^2)], \ x < \mu_{j1}; \\ 1, \ \mu_{j1} \le 1 \le \mu_{j2}; \\ \exp[-(x-\mu_{j2})^2/(2\sigma_{j2}{}^2)], \ x > \mu_{j2}. \end{cases}$$

And then, we use the Generalized KL information formula to optimize $T(\theta_j|x)$ to obtain $T^*(\theta_j|x)$, which will be smooth. If $S_2=S_1$, then $T^*(\theta_j|x)=P(x|y_j)/P(x)/\max[P(x|y_j)/P(x)]=T(\theta_2|x)$.

Figure 10 shows the MSI classification of ages for given prior distribution $P(x)$ and the truth functions of five labels. Five labels are ($y_1$, $y_2$, $y_3$, $y_4$, $y_5$)=("Adult", "Child", "Youth", "Middle age", "Old"). Figure 10 (a) shows the truth functions of five labels.

Among these truth functions, only each of $T(\theta_3|x)$ and $T(\theta_4|x)$ is constructed by two logistic functions; each of others is a logistic function.

We can also treat these truth functions as learning functions $P(\theta_j|x)$ obtained from the popular method, and then use the Bayes classifier or the Maximum Posterior Probability criterion to classify.

In Figure 10 (a), $P(x)$ is assumed: $P(x)=k[1-1/\exp[-0.1*(x-70)]$ for $x>0$, where $k$ is a normalizing constant. From the five labels, we can have compoud labels $y_0=y_1'$, $y_6=y_3\wedge y_1'$, and $y_7=y_3\wedge y_1$.

Figure 10 (b) shows the effect of the MSI classifier (see Eq. 3.12)).

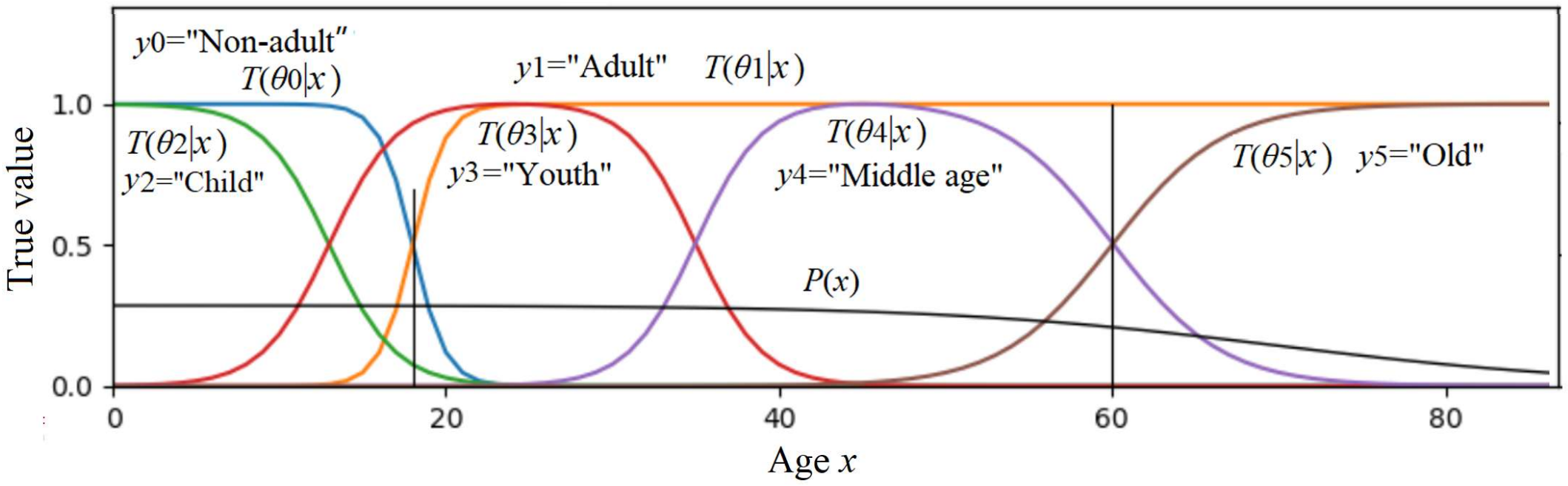

(a) The truth functions of five labels for ages and the prior distribution $P(x)$ of population.

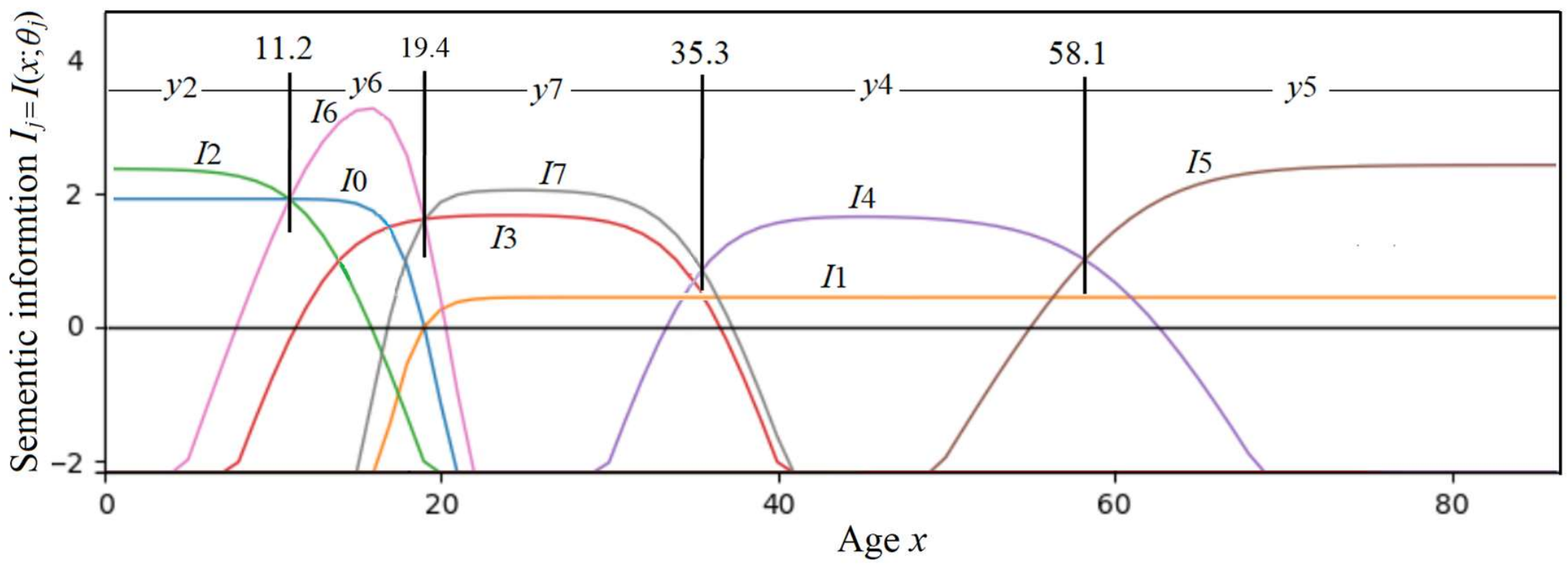

(b) Labeling $x$ according to which of $I_j=I(x;\theta_j)$ ($j$=0,1,…7) is miximum.

**Figure 10.** The maximum semantic information classification of ages.

Figure 10 indicates that the Maximum Posterior Probability (MPP) criterion and the MSI criterion result in different classifications. Using the MPP criterion, we shall only select $y_0$="Non-adult" or $y_1$="Adult" for most ages. However, using the MSI criterion, we shall select $y_2$, $y_6$, $y_7$, $y_4$, and $y_5$ in turn as age $x$ increases. The MSI criterion encourages us to use more labels with smaller logical probability. For example, if $x$ is between 11.2 and 16.6, we should use $y_6=y_3\wedge y_1'$="Youth" and "Non-adult". However, for most $x$, CM2 does not use redundant labels as Binary Relevance [52] does. For example, using the MSI criterion, we do not add label "Non-youth" to $x$=60 with label "Old" already.

### 4.2 The Results of CM3 for the MMI classifications of Unseen Instances

The author has tested CM3 by many examples.

**Example 4.2.1** The $z$ changes from 0 to 100; the step length is 1. Two Gaussian distributions $P(z|x_0)$ and $P(z|x_1)$ have parameters $\mu_0$=30, $\mu_1$=70, $\sigma_0$=15, and $\sigma_1$=10. $P(x_0)$=0.8 and $P(x_1$=0.2). The initial partitioning point $z'$ is 50.

**The iterative process**: Matching II-1 gets $z'$=53; Matching II-2 gets $z'$=54; Matching II-3 gets $z^*$=54.

The following is a two-dimensional example.

**Example 4.2.2 (**see Figure 11**)** There are three classes. The left two classes are two Gaussian distributions: $P(z|x_0)$ and $P(z|x_1)$; the right one is the mixture of two Gaussian distributions: $P(z|x_{21})$ and $P(z|x_{22})$. The sample size is 1000. See Table 3 for the parameters of four Gaussian distributions.

**Table 3.** The parameters of four Gaussian distributions

| | $\mu_{z1}$ | $\mu_{z2}$ | $\sigma_{z1}$ | $\sigma_{z2}$ | $\rho$ | $P(x_i)$ |
|---|---|---|---|---|---|---|
| $P(z\|x_0)$ | 50 | 50 | 75 | 200 | 50 | 0.2 |
| $P(z\|x_1)$ | 75 | 90 | 200 | 75 | -50 | 0.5 |
| $P(z\|x_{21})$ | 100 | 50 | 125 | 125 | 75 | 0.2 |
| $P(z\|x_{22})$ | 120 | 80 | 75 | 125 | 0 | 0.1 |

Two vertical lines make the initial partition.

**The iterative process:** It is shown in Figure 11.

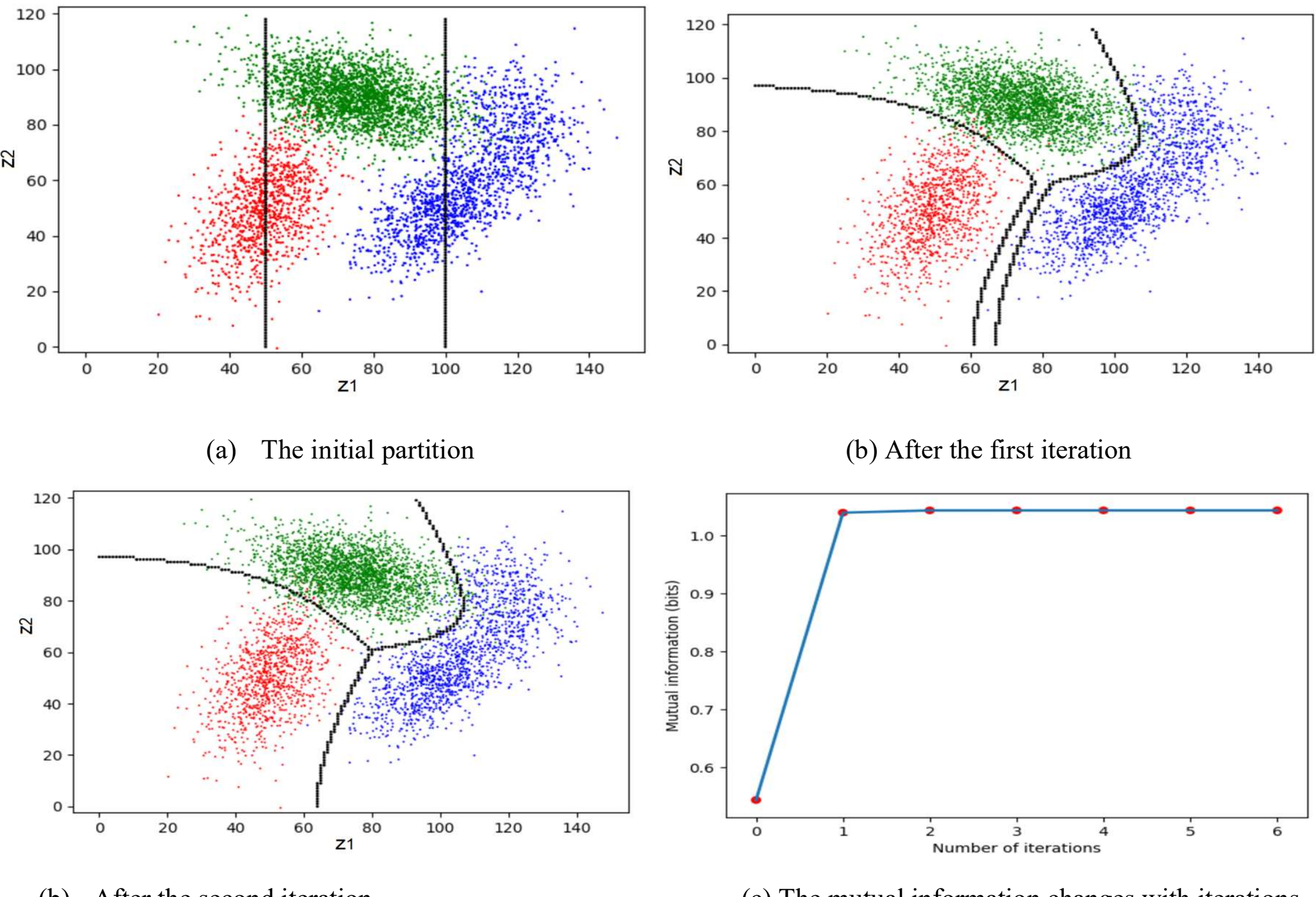

(a) The initial partition (b) After the first iteration

(b) After the second iteration (c) The mutual information changes with iterations

**Figure 11**. The MMI classification of unseen instances. The classifier is $y=f(z)$. The mutual information is $I(X;Y)$. $X$ is a true class and $Y$ is a selected label.

After two iterations, the mutual information $I(X; Y)$ is 1.0434 bits. The convergent MMI is 1.0435 bits. Two iterations make the mutual information reach 99.99% of the convergent MMI.

To test the reliability of CM3, the author used a very bad initial partition. The convergence is also very fast (see Figure 12).

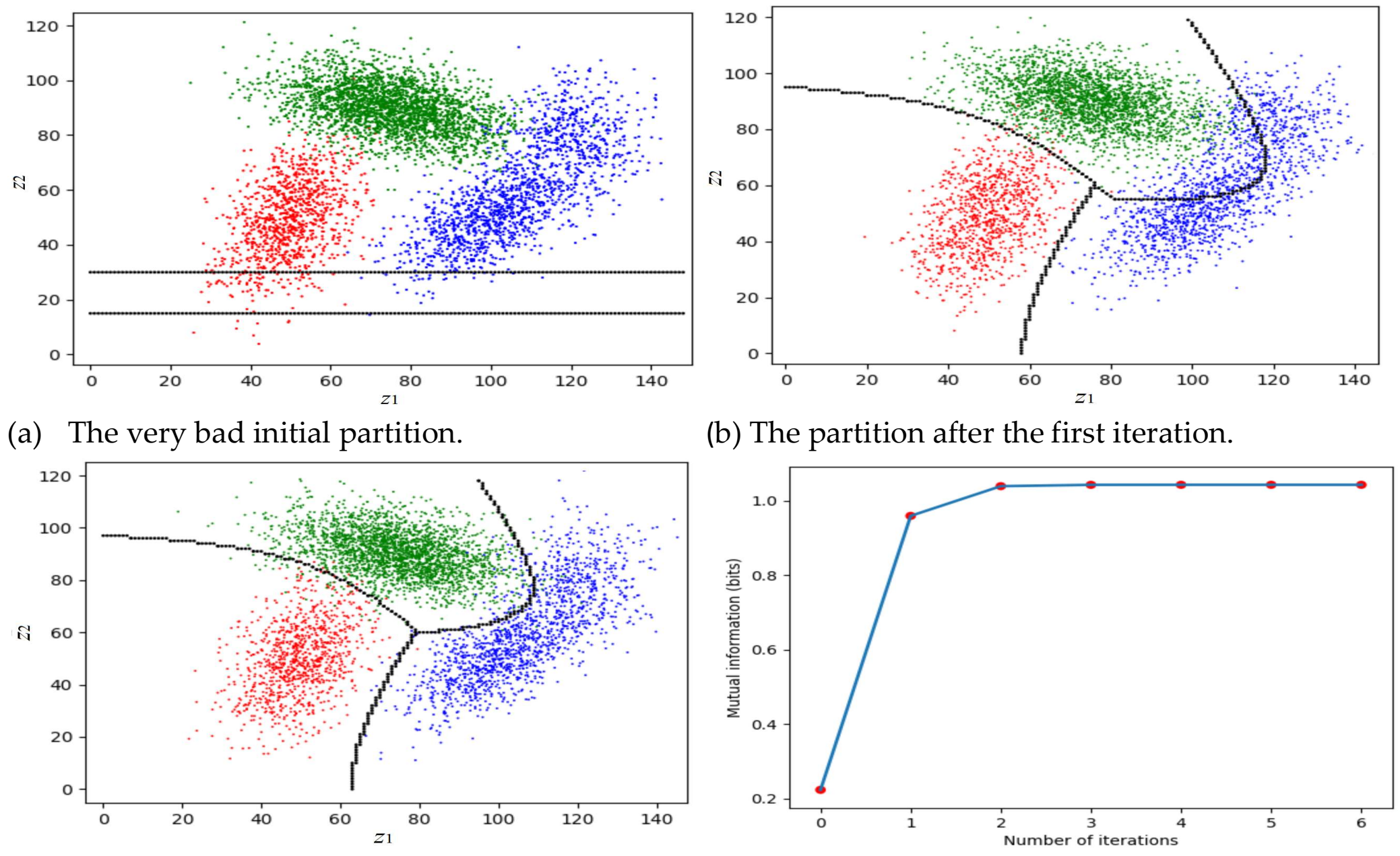

(a) The very bad initial partition. (b) The partition after the first iteration.

(c) The partition after the second iteration. (d) The mutual information changes with iterations

**Figure 12**. The MMI classification with a very bad initial partition.

The author has used the above example with different parameters and different initial partitions to test CM3. All iterative processes are fast and valid. In most cases, 2-3 iterations can make the mutual information surpass 99% of the MMI.

### 4.3. The Results of CM4 for Mixture Models

The following three examples show that the CM-EM algorithm can outperform the EM algorithm and the MM algorithm.

Ueda and Nakano [54] proposed an example to show that, because some initial parameters result in the local maximum of $Q$, local or invalid convergence is inevitable in the EM algorithm. This invalid convergence was also verified by Marin *et al.* [55]. The following is this example.

**Example 4.3.1**. A mixture model has two Gaussian components. The true model parameters are $(\mu_1^*, \mu_2^*, P^*(y_1), \sigma_1^*, \sigma_2^*) = (100, 125, 0.7, 10, 10)$. The invalid convergence center on the $\mu_1$-$\mu_2$ plane is $(\mu_1, \mu_2) = (115, 95)$, where $Q$ is a local maximum.

We use this example with initial parameters $(\mu_1, \mu_2, P(y_1), \sigma_1, \sigma_2) = (115, 95, 0.5, 10, 10)$ to test the CM-EM algorithm to see whether $(\mu_1, \mu_2,)$ can converge to $(\mu_1^*, \mu_2^*) = (100, 125)$. Figure 12 shows the result.

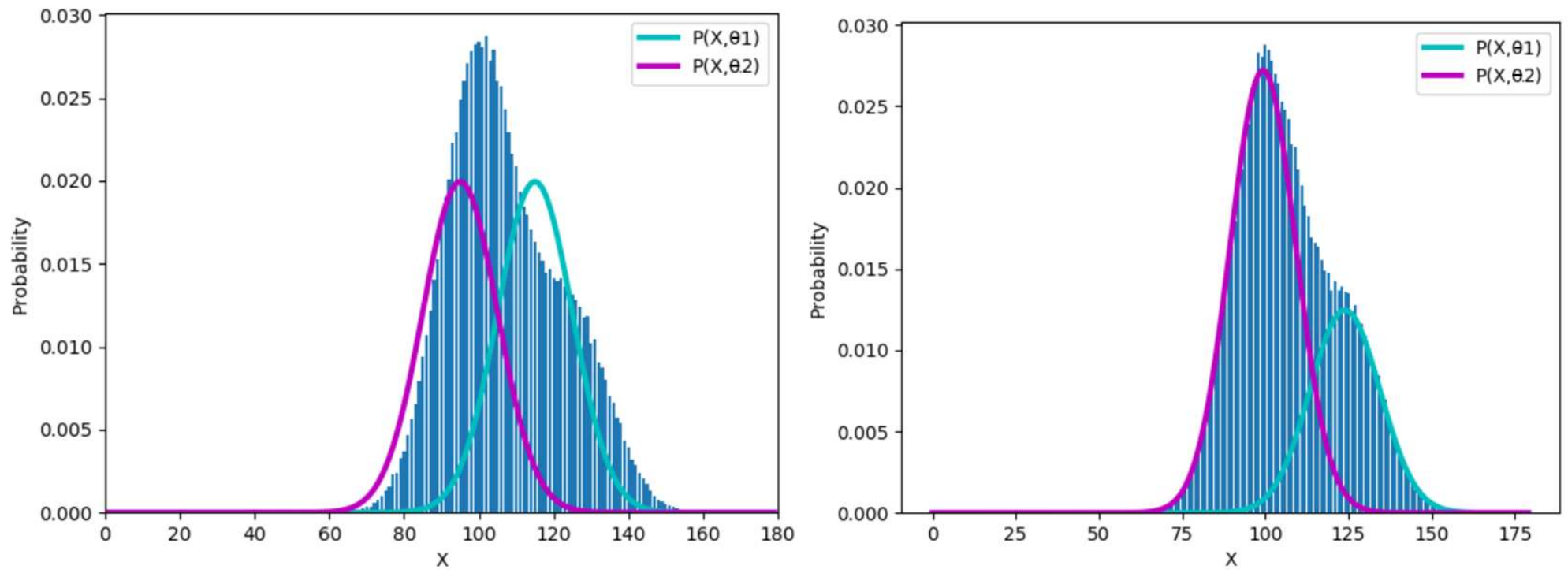

(a) Q is close to local maximum at beginning (b) $L_X(\theta)$ converges to the global maximum after 63 iterations

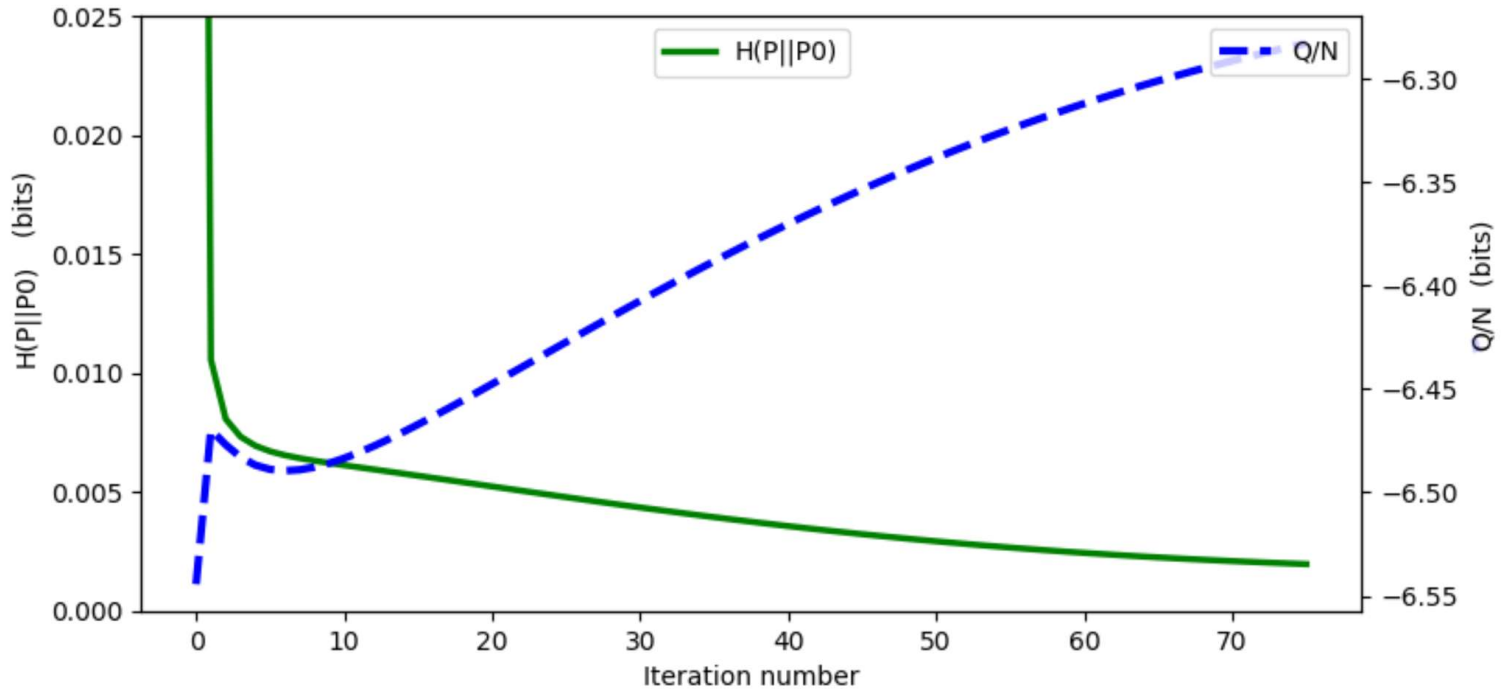

(c) $Q$ decreases after the first E2-step and then increases as $H(P||P_\theta)$ decreases in the CM-EM algorithm.

Figure 13. The iterative process from the local maximum of $Q$ to the global minimum of $H(P||P\theta)$. The stop condition is that the deviation of every parameter is less than 1%.

The following example is to compare the iteration numbers of different algorithms. Neal and Hinton [56] used this example to compare their Maximization-Maximization（MM) algorithm with the EM algorithm. Now we use the same example to compare the CM-EM algorithm with the EM algorithm and the MM algorithm.

**Example 4.3.2** Real and initial parameters including mixture ratios are shown in Table 4. The transforming formula is $x$=20($x$'-50), where $x'$ is an original data point and $x$ is a data point used for Table 4. Assume that $P(x)$ comes from the two Gaussian functions with real parameters. Using the CM-EM algorithm, we obtain $H(P||P_\theta)$=0.00072 bit after 9 E1 and E2-steps and 8 MG steps.

**Table 4. Trye** and guessed model parameters and iterative results of Example 4.3.2.

| | **Real Parameters** | | | **Starting parameters** $H(P\|\|P_\theta)$=0.68 bit | | | **Parameters after 9 E2-steps** $H(P\|\|P_\theta)$=0.00072 bit | | |
|---|---|---|---|---|---|---|---|---|---|
| | $\mu^*$ | $\sigma^*$ | $P^*(Y)$ | $\mu$ | $\sigma$ | $P(Y)$ | $\mu$ | $\sigma$ | $P(Y)$ |
| $y_1$ | 46 | 2 | 0.7 | 30 | 20 | 0.5 | 46.001 | 2.032 | 0.6990 |
| $y_2$ | 50 | 20 | 0.3 | 70 | 20 | 0.5 | 50.08 | 19.17 | 0.3010 |

The iterative process is shown in Figure 14.

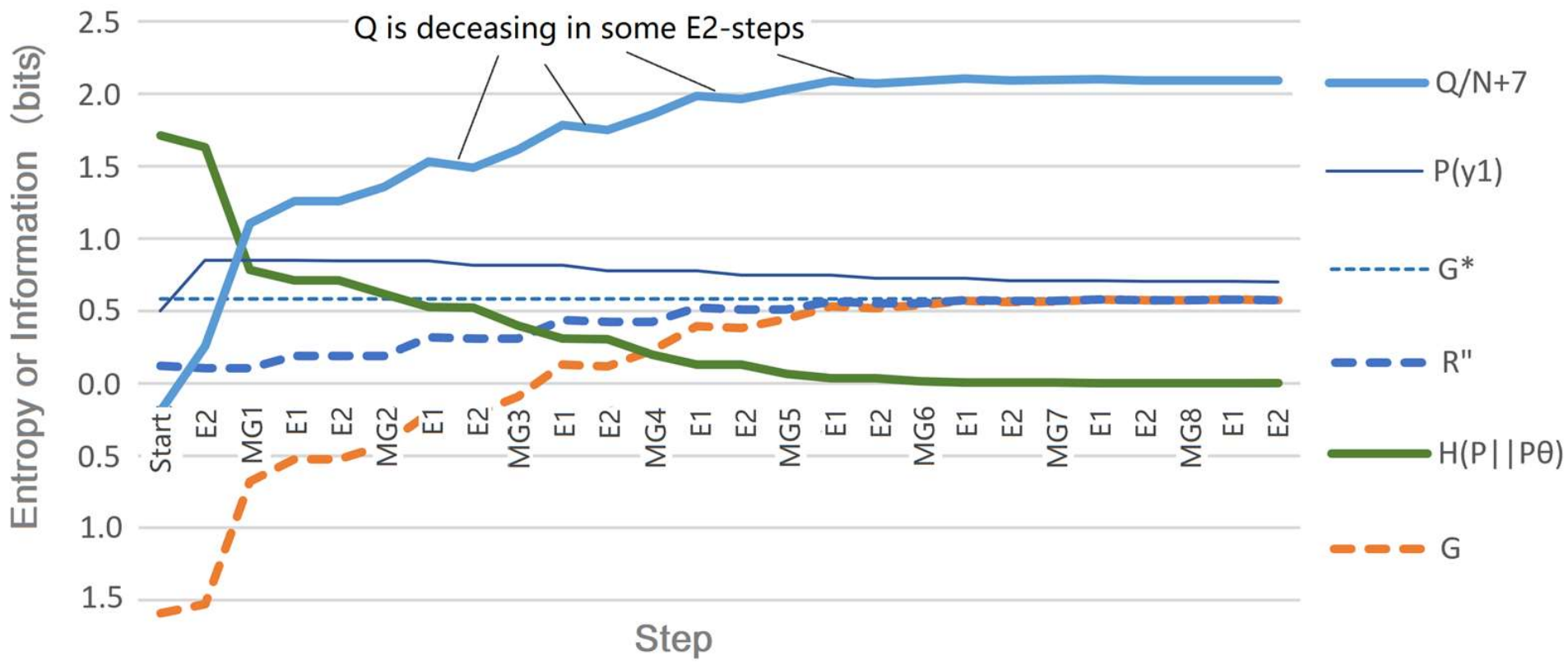

**Figure 14**. The iterative process of the CM-EM algorithm for Example 6.2. Some E2-steps decrease $Q$. The relative entropy is less than 0.001 bit after 8 iterations.

The author also used a sample whose size is 1000 to produce $P(x)$ to test the CM-EM algorithm. Table 5 shows the iteration numbers and final parameters for the three different algorithms.

**Table 5.** The iteration numbers and final parameters for different algorithms.

| Algorithm | Sample size | Iteration number | Convergent parameters | | | | |
|---|---|---|---|---|---|---|---|
| | | | $\mu_1$ | $\mu_2$ | $\sigma_1$ | $\sigma_2$ | $P(y_1)$ |
| EM | 1000 | about 36 | 46.14 | 49.68 | 1.90 | 19.18 | 0.731 |
| MM | 1000 | about 18 | 46.14 | 49.68 | 1.90 | 19.18 | 0.731 |
| CM-EM | 1000 | 8 | 46.01 | 49.53 | 2.08 | 21.13 | 0.705 |
| Real parameters | | | 46 | 50 | 2 | 20 | 0.7 |

These data show that iterations that the CM-EM needs are less than half of iterations that the EM or the MM algorithm needs.

**Example 4.3.3** There are six components in a two-dimensional feature space as shown in Figure 15. The sample size is 1000. True and initial parameters can be found through Appendix B.

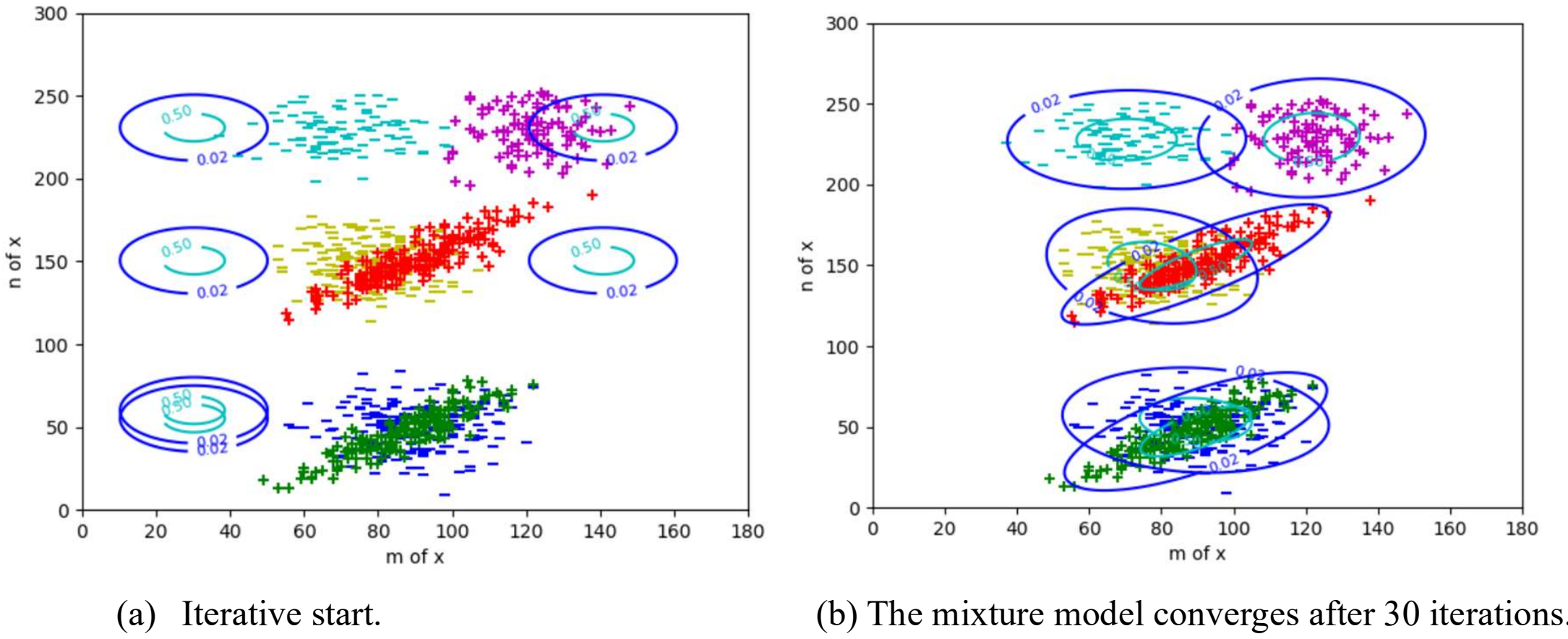

(a) Iterative start. (b) The mixture model converges after 30 iterations.

**Figure 15.** CM4 for a two-dimensional mixture model. There are six components with Gaussian distributions.

This example is to test if CM4 can validly converge for seriously overlapping components. The upper two pairs of components can quickly converge whereas the lower a pair can only slowly converge. The convergence condition is that horizontal deviation is smaller than 1.

## 5. Discussions

### 5.1 Discussing Confirmation Measure $b^*$

In modern times, the induction problem has become the confirmation problem [58]. There have existed many confirmation measures [59].

Most confirmation measures emphasize that larger $P(y_1|x_1)$ (more positive examples) is important whereas $b_1^*=b^*(y_1 \rightarrow x_1)$ emphasizes that smaller $P(y_1|x_0)$ (fewer negative examples) is important. For example, when sensitivity $P(y_1|x_1)$=0.1 and specificity $P(y_0|x_0)$=1, we can derive both $b_1^*$ and $CL_1$ are 1, which are reasonable. However, using the existing confirmation formulas [59], the degrees of confirmation of $MP_1$ are very small.

When sensitivity is 1, if specificity is small, such as 0.1, the degree of confirmation of $MP_1$, $b_1^*$, is also 0.1. However, using the existing confirmation formulas, the degrees of confirmation are much bigger than 0.1. This is unreasonable because the ratio of negative examples is 0.9/1.9≈0.47≈0.5, which means that $MP_1$ is almost unbelievable.

From the above two examples, we can find that confirmation measure $b^*$ emphasizes that no negative examples (for nonfuzzy hypotheses) or fewer negative examples (for fuzzy hypotheses) are more important than more positive examples, and hence, it is compatible with Popper's falsification thought [31,32]. This measure $b^*$ is compatible with confidence level and hence is also supported by medical practices.

Eells and Fitelson [60] suggested using Hypothesis Symmetry as a standard to evaluate various conformation measures. Hypothesis Symmetry means $b^*(y_1 \rightarrow x_1)= -b^*(y_1 \rightarrow \text{not } x_1)=-b^*(y_1 \rightarrow x_0)$. We can prove that confirmation measure $b^*$ has Hypothesis Symmetry:

$$b^*(y_1 \rightarrow x_1) = \frac{P(y_1|x_1)-P(y_1|x_0)}{\max(P(y_1|x_1),P(y_1|x_0))} = -\frac{P(y_1|x_0)-P(y_1|x_1)}{\max(P(y_1|x_0),P(y_1|x_1))} = -b^*(y_1 \rightarrow x_0). \quad (5.1)$$

### 5.2. Discussing CM2 for the Multilabel Classification

In comarision with popular methods, such as Binary Relevance [52], for Multilabel learning, CM3 does not need $n$ samples for $n$ pairs of labels. It can directly obtain the semantic channel that consists of a group of truth functions from the Shannon channel $P(Y|X)$ or the sampling distribution $P(x, y)$.

In comparison with the MPP criterion, the MSI criterion can reduce the rate of failing to report smaller probability events. When information is more important than correctness, the MSI criterion is better.

Note That the boundary for "Old" in Figure 10 (b) is not 60 but 58.1. It is because "Old" has smaller logical probability than "Middle age". If people's average lifetime is longer, the boundary for "Old" will move right. We can image that the new partitioning boundary will result in new sampling distribution $P(x|y_5)$ and new truth function $T(\theta_5|x)$; the new truth function will make the boundary move right further ⋯ The truth function or the semantic meaning of "Old" should have been evolving with human average lifetime in this way.

### 5.3 Discussing CM3 for the MMI Classifications of Unseen Instances

Solving MMI is a difficult problem not only with machine learning [19,20] but also with the classical information theory. Shannon and many researchers [7,6] use the least average distortion criterion instead of the MMI criterion to optimize detections and estimations. If we use the MMI criterion, the residual error coding will need smaller average code length. Why do not they use the MMI criterion? The reason is that it is hard to optimize partition boundaries for MMI. Now, using CM3, we can resolve this problem at least for low-dimensional feature spaces.

Popular methods for the MMI classifications or estimations use parameters to construct transition probability functions or likelihood functions and then optimize these parameters by the Gradient Descent or the Newton method. The optimized parameters ascertain partition boundaries. However, CM3 or the CM iteration algorithm separately constructs $n$ likelihood functions by parameters for $n$ different classes and then optimizes labels for different $z$. It provides the numerical solutions of partition boundaries. We compare CM3 and the Gradient Descent in Table 6.

**Table 6.** Comparing the CM algorithm and the Gradient Descent for low-dimensional feature spaces

| About | Gradient Descent | CM3 |
|---|---|---|
| Models for different classes | Optimized together | Optimized separately |
| Boundaries is expressed by | Functions with parameters | Numerical values |
| For complicated boundaries | Not easy | Easy |
| Consider gradient and search | Necessary | Unnecessary |
| Convergence | Not easy | Easy |
| Computation | Complicated | Simple |
| Iterations needed | Many | 2-3 |
| Samples required | Not necessarily big | Big enough |

The CM iteration algorithm also has two disadvantages. One is that it requires that every sub-sample for every class is big enough so that we can construct $n$ likelihood functions for $n$ classes. Another is that for high-dimensional feature spaces, it is not feasible to label every $z$. We need to combine the CM iteration algorithm with neural networks for the MMI classifications of high-dimensional feature spaces.

A neural network is a classifier $y=f(z)$. For a given neural network, Matching I is to let the semantic channel match the Shannon channel to obtain reward functions $I(X; \theta_j|z)$, $j$=0, 1,… For given reward functions, Matching II is to let the Shannon channel match the semantic channel to obtain new neural network parameters. Repeating two steps can make $I(X; \theta)$ converge to MMI. The Matching I and Matching II are similar to the tasks of generative and discriminative models in the the Generative Adversarial Network. Combining CM3 and popular deep learning methods [33], we should be able to improve the MMI classifications for high-dimensional feature spaces.

### 5.4. Discussing CM4 for mixture models

The above results in Section 4.3 indicate that the complete data log-likelihood $Q$ and the incomplete data log-likelihood $L_X(\theta)$ are not always positively correlated as most researchers believe. In some cases, $Q$ may and should decrease because $Q$ may be greater than $Q^*=Q(\theta^*)$, which is the true model's $Q$. In Example 3.4.1, assuming the true model's parameters $\sigma_1^*=\sigma_2^*=\sigma^*$ and $P^*(y_1)=P^*(y_2)=0.5$, we can prove that $P(y_1|x)$ and $P(y_2|x)$ are a pair of logistic functions and become sharp as $\sigma$ decreases. Hence, $H$ increases as $\sigma$ increases. We can prove partial derivative $\partial H/\partial\sigma^*>0$. Hence, when $\theta=\theta^*$,

$$\frac{\partial Q}{\partial \sigma}=\frac{\partial L_X(\theta)}{\partial \sigma}-\frac{\partial H}{\partial \sigma}=0-\frac{\partial H}{\partial \sigma}<0.$$

Therefore, we can find a small positive number $\Delta$ and replace $\sigma^*$ with $\sigma^*-\Delta$ so that $Q(\sigma^*-\Delta)>Q(\sigma^*)$.

The new convergence theory basing CM4 explains that CM4 can converge because the iteration can maximize $G/R$ or minimize $R-G$.

We have used different examples to test the CM-EM algorithm. The experiments show that the CM-EM algorithm can reduce slow and invalid convergence that the EM algorithm encounters when mixture ratios are imbalanced, or the local maximum $Q$ exists. In most cases, the CM-EM algorithm needs no more than ten iterations for the convergence of Gaussian mixture models. Its convergence speed is faster or similar to other improved EM algorithms, such as the MM algorithm [55] and the multiset EM algorithm [61].

The CM-EM algorithm can be used not only for Gaussian mixtures but also for other mixtures. For other mixtures, the MG step is a little difficult. The convergence proof should be the same.

CM4 and CM3 together can be used for unsupervised learning. For CM4, we can obtain a group of model parameters from a sample with distribution $P(x)$. Using CM3, we can make the MMI classification for the sample.

The CM-EM algorithm cannot avoid that $\theta$ converges to the boundary of parameter spaces. We need to combine some existing algorithms, such as, Split and Merge EM algorithm [62] and Competitive EM algorithm [63], for better global convergence of mixture models.

## 6. Conclusions

The semantic information G theory combines the thoughts of Shannon, Popper, Fisher, Zadeh, Carnap *et al*. The semantic information measure, the G measure, increases as the logical probability decreases as well as Carnap and Bar-Hillel's semantic information measure; however, the G measure also decreases as the relative deviation increases, and hence it can be used for hypothesis tests.

Logical Bayesian Inference (LBI) uses the truth function instead of the Bayesian posterior as the inference tool. Using the truth function $T(\theta_j|x)$, we can make probability predictions with the different prior $P(x)$, as using the Transition Probability Function (TPF) $P(y_j|x)$ or the Inverse Probability Function (IPF) $P(\theta_j|x)$. However, it is much easier to obtain the optimized truth function from samples than to obtain the optimized IPF because $P(y_j)$ or $P(\theta_j)$ is not necessary for optimizing truth functions. The more important is that the truth function can represent the semantic meaning of a hypothesis or a label and connect statistics and logic better. A windfall is that the optimization of the truth function brings a seemly reasonable confirmation measure for induction.

MultilabelMultilabelMultilabelA group of Channel Matching (CM) algorithms CM1, CM2, CM3, and CM4 are used to improve machine learning, especially to resolve the Multilabel-Learning-for-New-$P(x)$ problem. CM1 can be used to improve label learning and confirmation; CM2 can be used to improve Multilabel classifications; CM3 can be used to improve maximum mutual information classifications of unseen instances on low-dimensional feature spaces; CM4 can be used to improve mixture models. The *G* theory and LBI should have survived the tests by their applications to machine learning.

For the further applications of the *G* theory and LBI to machine learning, we need to combine the CM algorithms with the neural networks in the future. The logical Bayesian inference may be further developed for the unification of logic and statistics.

**Appendix A.** **Abbreviations**

| Abbreviation | Original text |
|---|---|
| BI | Bayesian Inference |
| CM | Channel Matching |
| CM-EM | Channel Matching Expectation-Maximization |
| EM | Expectation-Maximization |
| G theory | Semantic information G theory |
| GPS | Global Positioning System |
| HIV | Human Immunodeficiency Virus |
| IPF | Inverse Probability Function |
| KL | Kullback-Leibler |
| LBI | Logical Bayesian Inference |
| LI | Likelihood Inference |
| MLE | Maximum Likelihood Estimation |
| MM | Maximization -Maximization |
| MMI | Maximum Mutual Information |
| MPP | Maximum Posterior Probability |
| MSI | Maximum Semantic Information |
| MSIE | Maximum Semantic Information Estimation |
| SMI | Semantic Mutual Information |
| SHMI | Shannon's Mutual Information |
| TBP | Traditional Bayes' Prediction |
| TPF | Transition Probability Function |

## Appendix B Supplemental Materials

The supplemental materials **including several Python 3.6 source codes and a Excel file for Figures 9-15** can be downloaded from http://survivor99.com/lcg/cm/forGtheory.zip

**Table 1.** The list of Supplemental Materials

| **Program name** | **Task** |
|---|---|
| Bayes Theorem III 2.py (python 3.6 source file) | For Figure 7. To show Bayes theorem III formula 2 for membership function |
| Ages-MI-classification.py (python 3.6 source file) | For Figure 8. To show people classification on ages according to maximum semantic information criterion for given membership functions and $P(x)$. |
| MMI-v.py (python 3.6 source file) | For Figure 10. To show the Channels Matching (CM) algorithm for the maximum mutual information classifications of unseen instances. To modify parameters, see the program. The print window will display the change of the mutual information. When you run it, just close a window for the next window. |
| LocationTrap63.py (python 3.6 source file) | For Figure 11. To show how the CM-EM algorithm for mixture models avoids local convergence because of the local maximum of $Q$. |
| Folder DorEx6 (with Excel file and Word readme file) | For Figure 12. To show the effectiveness of every step of the CM-EM algorithm for the mixture model used by Neal and Hinton. |

**Author Contributions:**

Chenguang Lu is only author for all work including programming.

**Funding:**

This research received no external funding.

**Acknowledgments:** The author thanks Professor Peizhuang Wang for his long-term support and encouragement. The author also thanks the anonymous reviewers for their suggestions.

**Conflicts of Interest:** The author declares no conflict of interest.

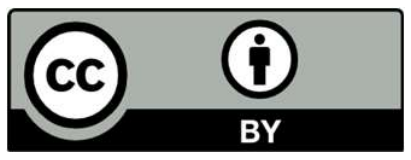